\documentclass[final,5p,times,twocolumn]{elsarticle}

\usepackage{amssymb}
\usepackage{amsfonts}
\usepackage{graphicx}
\usepackage{stfloats}
\usepackage{caption}
\usepackage{color}
\usepackage{float}
\usepackage{threeparttable}
\usepackage{multirow}
\usepackage{booktabs}
\usepackage{adjustbox}
\usepackage{flushend,cuted}
\usepackage{wrapfig} 
\biboptions{numbers,sort&compress} 

\usepackage{tikz}
\usepackage{comment}
\usepackage{amsmath,amssymb} 
\usepackage{color}

\usepackage{microtype}
\usepackage{bm}
\usepackage{mathtools} 

\usepackage{latexsym}
\usepackage{mathtools} 

\usepackage[ruled,vlined]{algorithm2e}
\usepackage{float} 
\usepackage{caption}
\usepackage{subcaption}

\def\argmax{\mathop{\rm arg\,max}\limits}
\def\argmin{\mathop{\rm arg\,min}\limits}

\def\rev#1{\textcolor{black}{#1}}

\journal{}

\begin{document}
\date{}
\begin{frontmatter}



\title{Reversing Skin Cancer Adversarial Examples by Multiscale \\ Diffusive and Denoising Aggregation Mechanism}

\author[contact1]{Yongwei Wang}
\ead{yongwei.wang@zju.edu.cn}
\author[contact2]{Yuan Li}
\ead{liyuan91700@163.com }
\author[contact3]{Zhiqi Shen}
\ead{zqshen@ntu.edu.sg}
\author[contact4]{Yuhui Qiao \corref{cor1}}
\ead{yh.qiao19@sjtu.edu.cn}

\cortext[cor1]{Corresponding author.}

\address[contact1]{Shanghai Institute for Advanced Study, Zhejiang University, Shanghai 201203, China\\}
\address[contact2]{Department of Critical Care Medicine, Qilu Hospital of Shandong University, Jinan, Shandong 250012, China\\}
\address[contact3]{Joint NTU-UBC Research Centre Of Excellence In Active Living For The Elderly, NTU, 50 Nanyang Avenue, 639798, Singapore\\}
\address[contact4]{Department of Pathophysiology, Shanghai Jiao Tong University School of Medicine (SJTU-SM), Shanghai 20025, China\\}

\begin{abstract}
Reliable skin cancer diagnosis models play an essential role in early screening and medical intervention. Prevailing computer-aided skin cancer classification systems employ deep learning approaches. However, recent studies reveal their extreme vulnerability to adversarial attacks --- often imperceptible perturbations to significantly reduce the performances of skin cancer diagnosis models. To mitigate these threats, this work presents a simple, effective, and resource-efficient defense framework by reverse engineering adversarial perturbations in skin cancer images. Specifically, a multiscale image pyramid is first established to better preserve discriminative structures in the medical imaging domain. To neutralize adversarial effects, skin images at different scales are then progressively diffused by injecting isotropic Gaussian noises to move the adversarial examples to the clean image manifold. Crucially, to further reverse adversarial noises and suppress redundant injected noises, a novel multiscale denoising mechanism is carefully designed that aggregates image information from neighboring scales. We evaluated the defensive effectiveness of our method on ISIC 2019, a largest skin cancer multiclass classification dataset. Experimental results demonstrate that the proposed method can successfully reverse adversarial perturbations from different attacks and significantly outperform some state-of-the-art methods in defending skin cancer diagnosis models. 
\end{abstract}



\begin{keyword}
robust skin cancer diagnostic model \sep reverse adversarial examples \sep adversarial defenses \sep multiscale diffusive process \sep denoising aggregation.


\end{keyword}

\end{frontmatter}

\section{Introduction}
In the artificial intelligence (AI) era, deep convolutional neural networks (DCNNs) have achieved remarkable performances in a variety of general vision tasks, e.g., imagery classification \cite{AlexNet, ResNet, dosovitskiy2020image}, object detection \cite{ren2015faster, redmon2016you}, and inverse problems \cite{lucas2018using, Revhashnet}. Meanwhile, DCNNs are gaining increasing attention in computer-aided diagnosis (CAD) for skin cancer recognition and processing, a crucial area closely related to human health and welfare. A skin cancer classification model can significantly improve the efficiency of early diagnosing skin cancers, a common type of disease for elderly people globally \cite{esteva2017dermatologist, wang2022ssd}. Early detection and screening make it possible for medical intervention and successful treatment. Notably, a well-trained DCNN was demonstrated to reach expert-level capability in skin lesion classification in both binary and multiclass scenarios \cite{esteva2017dermatologist}. Consequently, the U.S. Food and Drug Administration (FDA) approved the first AI-driven medical diagnostic system in 2018 \cite{finlayson2019adversarial}.

\begin{figure}
\centering
\includegraphics[width=0.48\textwidth]{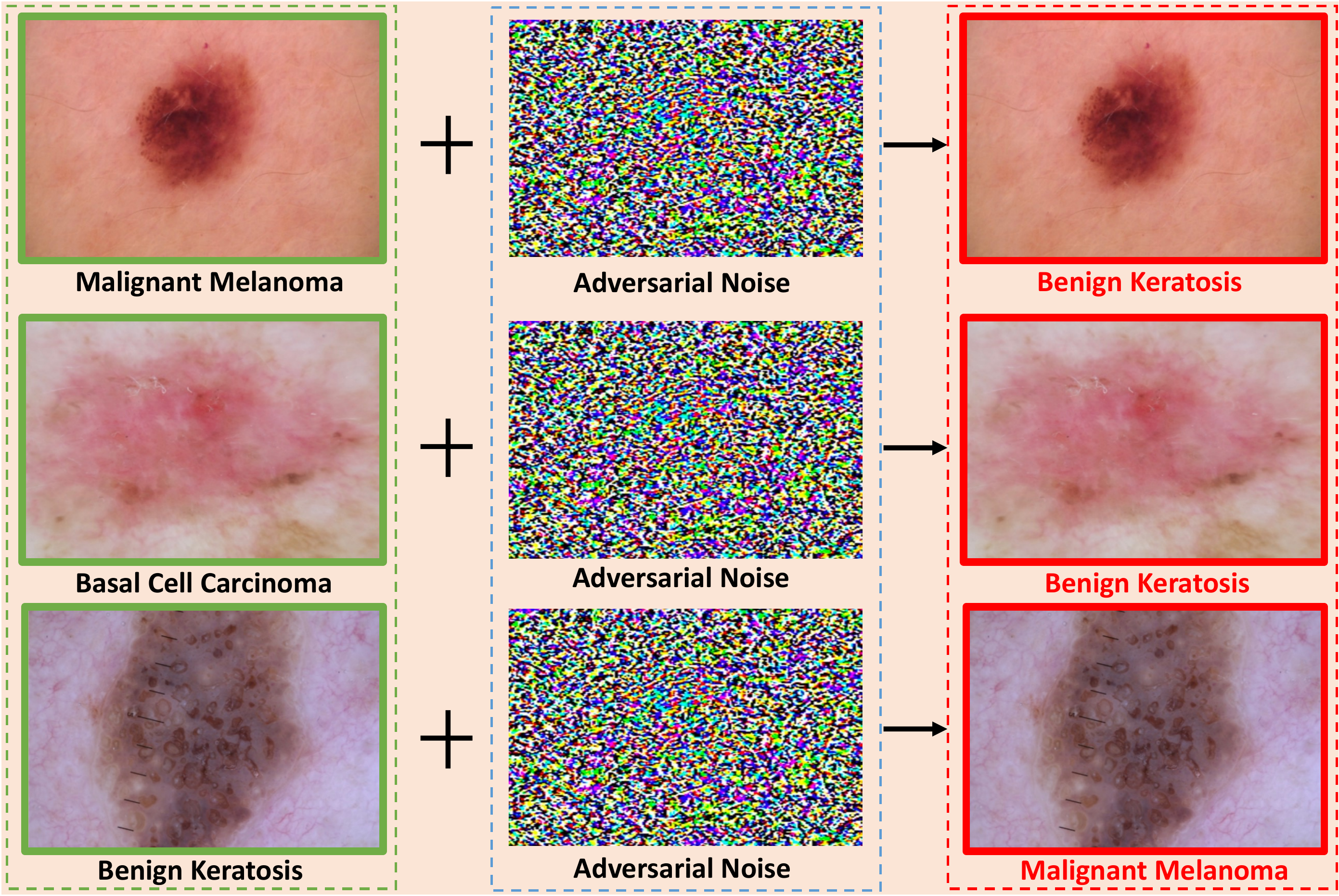}
\caption{ Illustration of adversarial attacks in skin cancer diagnosis models. The left column displays three typical clean skin images with their labels respectively as Malignant Melanoma (cancer), Basal Cell Carcinoma (cancer), and Benign Keratosis (benign). These images can be correctly recognized by a medical diagnostic model. By injecting slight adversarial noises (shown in the middle column) into these images \cite{FGSM}, we obtain adversarial skin images (shown in the right column) that can easily mislead medical diagnosis models: e.g., cancer images will be recognized as benign images. }
\label{fig:adv_example}
\end{figure}

Despite the great potential in facilitating automatic diagnosis and decision support, DCNN-based skin lesion recognition models are shown extremely vulnerable to adversarial attacks \cite{finlayson2019adversarial, ma2021understanding, kaviani2022adversarial}. In an adversarial attack process, an attacker may intentionally manipulate a tested image slightly, yet he/she can change the model predictions significantly \cite{FGSM, MIFGSM, ding2021towards, wang2021perception}. The manipulated images are known as \textit{adversarial examples}, which often appear visually imperceptible with clean images. As illustrated in Fig.~\ref{fig:adv_example}, adversarial skin examples (right column) can easily trick a medical diagnostic model by injecting some slight and human indecipherable noises (middle column) into clean images (left column). Indeed, the existence of adversarial skin examples can pose huge threats to automatic skin cancer diagnosis in practical scenarios. For example, a patient diagnosed with skin cancer may first manipulate the skin images using adversarial attacks, then send these images to an insurance company. The insurance company that relies on DCNN-based skin cancer diagnosis models can be misled to make wrong decisions and suffer from financial losses (e.g., approving reimbursements) \cite{finlayson2019adversarial, kaviani2022adversarial}.

In DCNN-based medical diagnosis systems, in particular, it is crucial to investigate effective defenses to protect victim diagnostic models from adversarial examples. Ideally, we have three requirements to evaluate a defense method: 

\begin{enumerate}
    \item \textbf{Attack-agnostic} property. Given the availability of diverse attack methods \cite{FGSM,MIFGSM,Iteradv17,CW17}, a defense is expected to deal with different attacks and even unseen ones. 
    
    \item \textbf{Training-free} property. Since re-training a diagnostic model takes much time and can be prone to overfitting issues, a defense mechanism free from training is desired that can work across different neural network architectures \cite{Xu0Q18, ding2021delving}. 
    
    \item \textbf{Resource-efficient} property. Due to the deployment of portable skin diagnosis models (e.g., \cite{wang2022ssd}), a resource-efficient defensive approach is imperative that can work on deployed models on mobile devices. 
\end{enumerate}

Existing adversarial defenses mainly focus on protecting natural images \cite{FGSM,madry2018towards,cohen2019certified, li2019certified,mustafa2019image, naseer2020self,ding2021delving, nie2022diffusion}. There are two mainstream research directions among these defenses. The first stream is a robustness-based defense that attempts to robustify untrained classifiers by augmenting clean images with their adversarial counterparts, e.g., \cite{FGSM, madry2018towards}. However, these methods often necessitate intensive training and they are often not scalable to large datasets. Some other studies try to certify a pretrained model by computing a perturbation radius within which this model can be shown provably robust to adversarial attacks \cite{cohen2019certified, li2019certified, xu2022medrdf, carlini2022certified}. Unfortunately, the obtained radius is often too small and they fail to defend against color-based attacks \cite{wang2021perception}. The other mainstream research resorts to purifying adversarial perturbations from adversarial examples \cite{Xu0Q18,mustafa2019image,naseer2020self,ding2021delving,nie2022diffusion}. However, our experiments on skin images reveal that these typical methods are either lack defending effectiveness (\cite{Xu0Q18,mustafa2019image}), or too resource-consuming to be deployed in edge devices. For example, works \cite{ding2021delving,nie2022diffusion} require the utilization of a graphics processing unit (GPU) and they take a few minutes to process a single image on an advanced GPU server. The intensive requirement for computing hinders their practical applications.  

To resolve these challenges above, this work proposes a multiscale diffusive and denoising aggregation (MDDA) framework to reverse skin adversarial examples. Different from defenses that explicitly utilize the knowledge of victim classifiers (e.g., \cite{ding2021delving, xu2022medrdf}), our defense is model-agnostic and has no assumptions on access to model decisions or model architectures. Thus our method is more practical with broader applications. Since medical diagnostic models often utilize image structures (e.g., texture information) to make predictions \cite{ma2021understanding, wang2022ssd}, we leverage a multiscale analysis to preserve key distinguishable features during the reverse engineering process. 

\begin{figure}
\centering
\includegraphics[width=0.48\textwidth]{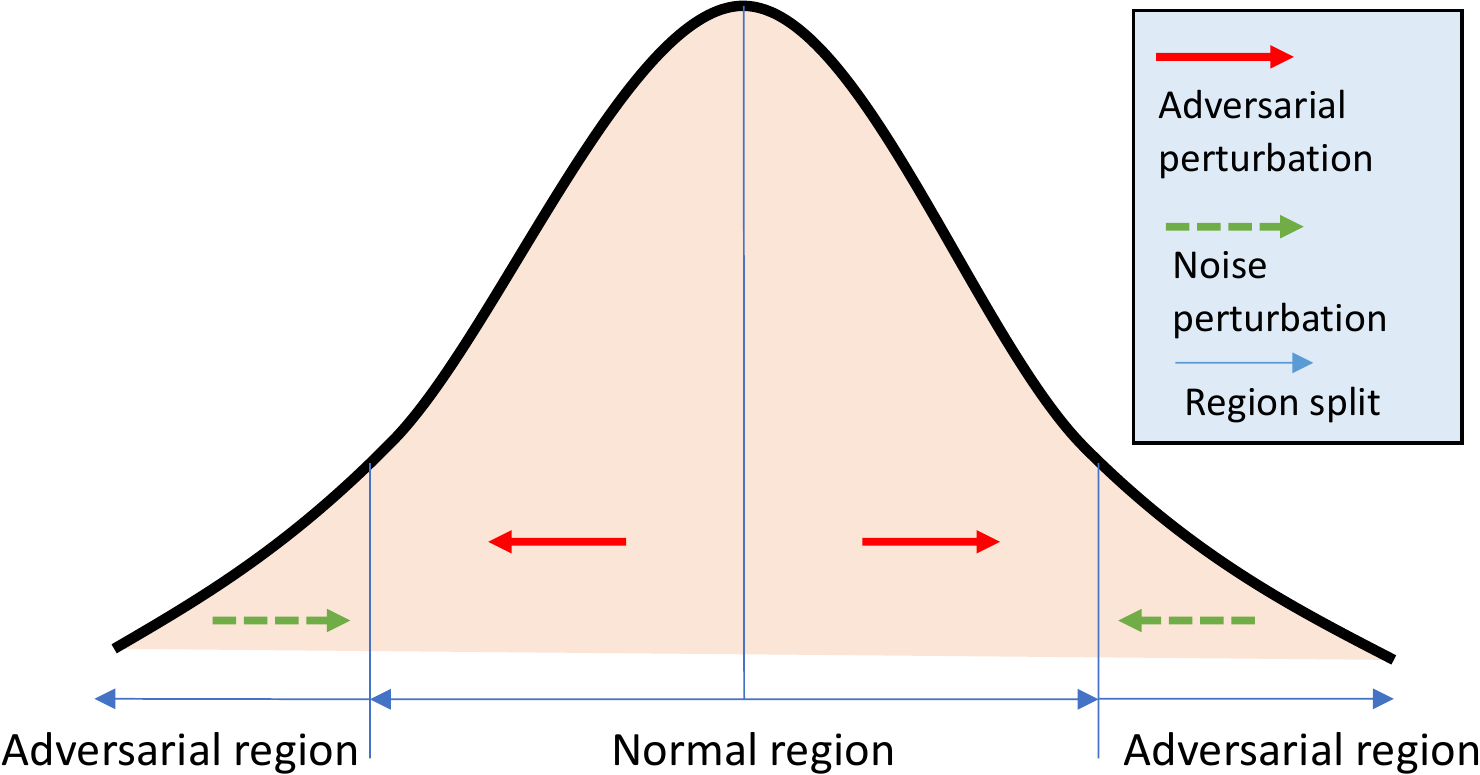}
\caption{Illustration of the ``fight-fire-with-fire'' mechanism to reverse skin adversarial examples. During adversarial attacks, images injected with adversarial noises will move from normal to adversarial regions. In the reverse process, we inject isotropic noises into manipulated samples which will ``drag'' these samples from low-probability regions back to normal ones. }
\label{fig:adv_distribution}
\end{figure}

A popular hypothesis for adversarial examples is that these samples take a much lower probability than those normal samples. As illustrated in Fig.~\ref{fig:adv_distribution}, normal images (i.e., normal regions) take up much space in the data distribution, while adversarial regions lie in the tail region. By perturbing with adversarial noises, clean images from normal regions will ``fall down'' to the adversarial region (indicated in red arrow). To reverse the attacking process, an intuitive approach is to ``drag'' the maliciously manipulated samples back to the normal region. However, it is challenging to achieve this purpose mainly for two reasons. Firstly, since a defender is unaware of the exact directions of adversarial noises and these noises show different directions for different samples, it is practically infeasible to estimate adversarial noises and reverse them. Secondly, the magnitudes of injected adversarial noises are also opaque to the defender. Thus it is difficult to determine the amount (or magnitude) of signals, i.e., reversing signals, to be injected into adversarial examples during the reverse process. A too-small magnitude of a reversing signal may not be enough to restore adversarial examples, yet a too-large one may damage the key information in the original sample.

Inspired by certified defenses \cite{cohen2019certified, li2019certified}, we tackle the reversing process by diffusing Gaussian isotropic noises to adversarial examples. \rev{Due to the isotropic property, the Gaussian noise can cover different regions and directions. Thus the injected Gaussian noise is likely to move perturbed images from the low-probability adversarial region back to the high-probability normal regions which is hypothesized to destroy the structure of adversarial noises.} However, different from \cite{cohen2019certified, li2019certified}, we leverage a multiscale processing technique, and more importantly, we propose to inject noises followed by a denoising module in a progressive manner. In general, a denoising module cannot effectively deal with heavy noises. Thus we propose to split injecting heavy noises into separate iterations. At each iteration, we inject a small number of noises into skin adversarial examples to disrupt adversarial noises in certain directions. We then denoise redundant noises utilizing the denoising module. To make better use of multiscale analysis, we propose to fuse image information by aggregating subimages at neighboring scales. After repeating the diffusion-denoise-aggregation procedure several times, adversarial images are likely to be ``dragged'' from adversarial regions back to the normal regions. Finally, we fuse subimages at different scales to obtain the reversed images.

In summary, our main contributions can be summarized as follows:

\begin{enumerate}
    \item We propose a novel framework to reverse skin adversarial examples to normal samples. We leverage a multiscale analysis to fit skin image processing. To tackle unknown directions and magnitudes of adversarial noises, we develop a diffusive and denoising aggregation block that can gradually move adversarial examples to normal regions. 
    
    \item The proposed method is attack-agnostic and model-agnostic. Thus it can protect skin diagnostic models from different even unseen attacks and our method is suitable for cross-model protection. 
    
    \item We conducted defensive experiments on ISIC 2019, a largest skin cancer classification dataset. Experimental results reveal the defense effectiveness and efficiency of the proposed method over state-of-the-art baseline methods.

\end{enumerate}

\section{Mathematical Model of Adversarial Attacks}
\label{sec:attacks}

In this section, we present the mathematical model of adversarial attacks. Assume that we have a well pretrained skin cancer diagnostic model:  $f:\mathcal{D} \subseteq \mathbb{R}^d \mapsto \mathbb{R}^K$, where $\mathcal{D}=[0, 255]^d$. Given a skin image $\boldsymbol{x} \in \mathbb{R}^d$, the diagnostic model can correctly predict the image's label as $y \in \mathcal{Y}$, i.e., $y=\argmax_{k=1,\cdots, K} f_k(\boldsymbol{x})$.

In the adversarial attack problem, an attacker attempts to find an $\epsilon$-ball bounded perturbation $||\boldsymbol{\delta}||_p \leq \epsilon$ within the vicinity of an image $\boldsymbol{x}$. Here $\epsilon \in [0, 255]$ denotes a perturbation budget, $||\cdot||_p$ denotes the $\ell_p$ norm constraint, and $p$ is often set as $\infty$. A skin image injected by a perturbation $\boldsymbol{\delta}$ will mislead the pretrained diagnostic model,

\begin{equation}
    \argmax_{k=1,\cdots,K} \; f_k(\boldsymbol{x} + \boldsymbol{\delta}) \neq y, \quad ||\boldsymbol{\delta}||_p \leq \epsilon \quad \textrm{and}\quad \boldsymbol{x} + \boldsymbol{\delta} \in \mathcal{D} 
\end{equation}

Suppose an attacker adopts a surrogate function $\mathcal{L}$, then the constrained optimization problem can be formulated as,

\begin{equation}
    \argmax_{\boldsymbol{\delta}} \; \mathcal{L}(f(\boldsymbol{x}+\boldsymbol{\delta}), y) \quad \textrm{s.t.} \; ||\boldsymbol{\delta}||_p \leq \epsilon, \; \boldsymbol{x} + \boldsymbol{\delta} \in \mathcal{D}
    \label{eq:obj_fun}
\end{equation}
where $\mathcal{L}$ is selected as the cross-entropy function following existing literature \cite{FGSM, wang2021perception}. 

Assume an attacker has white-box access to a victim diagnostic model (a.k.a white-box attack), then he/she can utilize gradient-based attacks to generate adversarial perturbations. Goodfellow et al \cite{FGSM} proposed the fast sign gradient method (FGSM) which computes the gradient of $\mathcal{L}(f(\boldsymbol{x}+\boldsymbol{\delta}), y)$ with respect to $\boldsymbol{x}$,

\begin{equation}
    \boldsymbol{\delta}_{FGSM} = \epsilon \cdot \textrm{sign}(\nabla_{\boldsymbol{x}} \mathcal{L} (f(\boldsymbol{x}+\boldsymbol{\delta}, y))
    \label{eq:fgsm}
\end{equation}
where $\boldsymbol{\delta}_{FGSM}$ denotes the adversarial perturbation from FGSM, $\textrm{sign}(\cdot)$ is an element-wise sign operator that produces $+1$ for positive gradients, $-1$ for negative gradients, and it gives $0$ for a zero gradient.

After obtaining perturbation $\boldsymbol{\delta}_{FGSM}$, an adversarial example $\boldsymbol{x}^{*}$ can be crafted using FGSM attack,
\begin{equation}
    \boldsymbol{x}^{*} = \boldsymbol{x} + \boldsymbol{\delta}_{FGSM}
\end{equation}

FGSM attack is a one-step gradient-based attack method. Its multi-step variants can be obtained by splitting the perturbation budget $\epsilon$ into $T$ quota where $T$ denotes the number of steps \cite{Iteradv17, MIFGSM, xie2019improving}. Besides gradient-based attacks, Carlini and Wagner proposed a regularization-based unconstrained optimization approach, named CW attack \cite{CW17}, to generate adversarial examples,
\begin{equation}
    \boldsymbol{x}^{*} = \argmin_{\boldsymbol{x}'} || \boldsymbol{x}' - \boldsymbol{x} ||^2 + \lambda \; \Bigg \{ \textrm{max} \bigg ( f_y (\boldsymbol{x}') - \textrm{min}_{j \neq y }  \Big \{ f_j( \boldsymbol{x}') \Big \}  \bigg ), \; 0 \Bigg \}
    \label{eq:cw}
\end{equation}
where $\lambda$ denotes a regularizer for the misclassification term. The CW attack is widely used to evaluate the robustness of a model in the white-box setting because it can achieve high attack success rates while only requiring minimum adversarial perturbations. To conduct CW attacks on skin diagnostic models, we adopt the default $\lambda$ as indicated in \cite{CW17}.

\section{Method of Reversing Skin Adversarial Examples}
\label{sec:proposed_method}

\begin{figure*}
\centering
\includegraphics[width=0.95\textwidth]{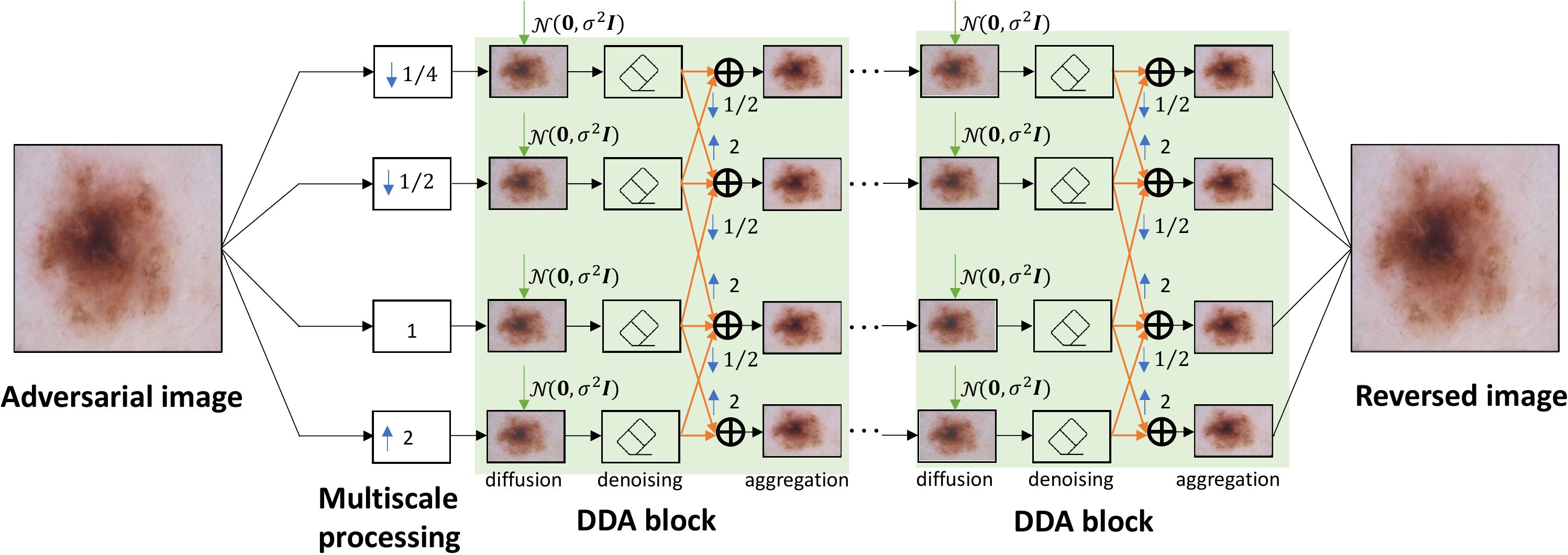}
\caption{\rev{Illustration of the framework of the proposed method. In the first stage, an adversarial image is processed with multiscale analysis: the image will be downsampled by a factor of 1/2 and 1/4, respectively, and upsampled by a factor of 2.  Then in the second stage, we design and insert $N$ diffusive and denoising aggregation mechanism (DDA) blocks sequentially. Each DDA block involves a diffusive process (Section 3.2), a denoising process (Section 3.3), and an aggregation process (Section 3.4). The output samples from the last DDA block will be inversely processed to the original scale and smoothed to obtain the reversed image.}  }
\label{fig:framework}
\end{figure*}

In this section, we describe in detail the procedure of the proposed multiscale diffusive and denoising aggregation (MDDA) method. \rev{The framework of MDDA is shown in Fig.~\ref{fig:framework} utilizing multiscale analysis and a sequential of diffusive and denoising aggregation (DDA) blocks.} The input of our method is a skin adversarial image and the output is a reversed image that removes adversarial perturbations from the input image. Please note that our method can also support an input as a clean image, in case we cannot distinguish adversarial examples from clean ones in advance. \rev{The MDDA framework consists of four key components: multiscale processing (Section 3.1), diffusive process (Section 3.2), denoising process (Section 3.3), and aggregation process (Section 3.4).} We introduce each component individually in the following sections.

\subsection{Multiscale Processing}
Skin cancer image diagnosis relies on some low-level features, e.g., textures and shapes of lesion regions. Therefore, we propose to utilize the multiscale processing technique to preserve this structural information for reliable model prediction. 

Without loss of generalizability, assume an input (adversarial) image as $\boldsymbol{x}$, and a multiscale processing operator as $\mathcal{T}$ with a scale as $k \in \mathcal{K}$ where $\mathcal{K}$ denotes the set of image scales. Applying a transform operator $\mathcal{T}$ to image $\boldsymbol{x}$, we can obtain a processed image $\boldsymbol{x}_{(k)}$ at scale $k$,
\begin{equation}
    \boldsymbol{x}_{(k)} = \mathcal{T} (\boldsymbol{x}, k)
    \label{eq:mulscale_opt}
\end{equation}
where the scale $k$ can be $2^{-n}$, \rev{$n \in \mathbb{Z}$ where $\mathbb{Z}$ denotes an integer set.} Given $\boldsymbol{x}_{(k)}$ at scale $k$, we can readily obtain its neighboring scale version as $\boldsymbol{x}_{(k')} = \mathcal{T} (\boldsymbol{x}, k'/k)$.

Similarly, we can define a reverse operator to approximately recover $\boldsymbol{x}$ from its scaled version $\boldsymbol{x}_{(k)}$,
\begin{equation}
    \boldsymbol{x} \approx \mathcal{T}^{-1} (\boldsymbol{x}_{(k)}, k )
    \label{eq:reverse_mulscale_opt}   
\end{equation}
where $\mathcal{T}^{-1}$ denotes a reverse operator. 

In general, an image pyramid formed by more scales can provide richer structural information, yet it brings about more computational costs. For example, a pyramid with $K$ scales from an image of a square size $M \times M$ has an overall number of pixels as $M^2 \cdot \sum_{k \in K} k $. Considering the computational requirement, as shown in Fig.~\ref{fig:framework}, we select four scales in our framework, i.e., $k=1/4, 1/2, 1, 2$, and we can achieve competitive performances. For downsampling scales (i.e., $1/4, 1/2$), we employ \textit{nearest neighbor} sampling and \textit{bicubic} sampling for the upsampling scale $2$.

\subsection{Diffusive Process} \label{sec:diffusion}
A diffusive process injects isotropic Gaussian noises iteratively into an image until it turns into an image that follows a Gaussian distribution \cite{ho2020denoising},
\begin{equation}
    \boldsymbol{x}^t = \alpha_t \boldsymbol{x}^{t-1} + \mathcal{N}(0, \beta_t I)
    \label{eq:diffusion}
\end{equation}
where $t$ denotes a diffusion step ($t=1, \cdots, T$ \rev{with $T$ denoting the maximum step}), $\alpha_t$ and $\beta_t$ denote the scaling parameter, and the variance of a Gaussian distribution at step $t$, respectively. As illustrated in Fig.~\ref{fig:adv_distribution}, the diffusive process is likely to destroy the structure of adversarial noises, thus it has the potential to reverse engineering the skin adversarial examples.   

Nevertheless, the diffusion model was primarily designed as a generative model and its reverse process is computationally demanding, \rev{i.e., it requires hundreds to thousands of step iterations on a GPU server due to the utilization of a corresponding $T$ value (e.g., $T=1000$ \cite{ho2020denoising}) In our defense method, however, the ``resource-efficient'' property requires a relatively small $T$. Indeed, we observe that diffusing samples with a few steps appears sufficient to ``drag'' samples from adversarial regions to normal ones in experiments. In the diffusive process, therefore, we only need to truncate the diffusive process by utilizing the first few $T^{'}$ ($T^{'} \ll T$) diffusion steps}. Intuitively, the reverse process of a truncated diffusive process is to employ a carefully-selected training-free function to denoise redundant Gaussian noises from $\boldsymbol{x}^{T'}$. However, our preliminary experiments indicate that non-training denoising functions cannot effectively deal with images with heavily perturbed noises. Thus, instead of denoising after diffusion for $T^{'}$ steps, we denoise the noise-injected image after each diffusion step. 

For simplicity, we select $\alpha_t$ as 1 and $\beta_t$ as $\sigma^2/T^{'}$ where $\sigma^2$ denotes the variance of a Gaussian distribution. Denote a denoising operator as $\mathcal{C}$, and denote a projection operator as $\mathcal{P}$ that projects samples to the valid image space, then we obtain an updating rule of the diffusion-denoising mechanism at step $t$ for scale $k$,
\begin{equation}
    \boldsymbol{x}_{(k)}^t = \mathcal{P} \circ  \mathcal{C} \bigg ( \boldsymbol{x}_{(k)}^{t-1} + \mathcal{N} \Big( 0, \; \sigma^2/T^{'} \Big)  \bigg )
    \label{eq:diff_denoise}
\end{equation}
where $\boldsymbol{x}_{(k)}^0 \triangleq \mathcal{T} (\boldsymbol{x}, k), t=1,\cdots,T^{'}$, and $\circ$ denotes a composite function that applies a denoising operator $\mathcal{C}$ followed by a projection operator $\mathcal{P}$ \rev{that clips image values $x$ element-wisely to the valid image space [0, 255], i.e., $\mathcal{P}(x) \triangleq
\textrm{min} ( \textrm{max} (x, 0), 255 ) $}. We will introduce the selected denoising function in the following section.

\subsection{Denoising Process} \label{sec:denoise}
The selection of a denoising function is crucial in the diffusion-denoising mechansim in Eq.~\ref{eq:diff_denoise}. The function should be capable of effectively purifying additive Gaussian noises from given noisy images even at low signal-to-noise ratios. Besides, the denoising process should well maintain the structure (e.g., edges, textures) of an input image, which are key features to make a reliable decision. 

To denoise white Gaussian noise, a total variation regularization method was developed in \cite{rudin1992nonlinear}. This model is also known as the Rudin, Osher, and Fatemi (ROF) model. The ROF model aims to find a denoised image $ \boldsymbol{x}'$ that satisfies an unconstraint optimization problem,
\begin{equation}
    \argmin_{\boldsymbol{x}' \in BV(\Omega)} || \boldsymbol{x}' ||_{TV(\Omega)} + \frac{\gamma}{2} \int_{\Omega} \Big ( \boldsymbol{x}' (u) - \boldsymbol{x} (u) \Big )^2 du
    \label{eq:tv_continuous}
\end{equation}
where $BV(\Omega)$ denotes a bounded variation space, $||\cdot||_{TV(\Omega)}$ denotes a total variation seminorm, and $\lambda$ is a regularizer.  

Assume $\boldsymbol{x}'$ being a smooth signal, then its total variation seminorm can be expressed as,
\begin{equation}
    || \boldsymbol{x}' ||_{TV(\Omega)} = \int_{\Omega} |\nabla \boldsymbol{x}' | du
    \label{eq:tv_norm}
\end{equation}
where $\nabla \boldsymbol{x}'$ denotes the gradient of image $ \boldsymbol{x}'$. The penalty term in Eq.~\ref{eq:tv_norm} attempts to minimize spurious details from a noisy image by discouraging the existence of oscillations. 

For the discrete 2D image signal $\boldsymbol{x}' \in \mathbb{R}^{M \times M}$, the total variation seminorm can be approximated by horizontal and vertical derivatives \rev{\cite{chambolle1997image, chambolle2010introduction}},

\begin{equation}
    || \boldsymbol{x}' ||_{TV(\Omega)} \approx \sum_{i=0}^{M-1} \sum_{j=0}^{M-1}  | \boldsymbol{x}'_{i+1,j} - \boldsymbol{x}'_{i,j} | + |\boldsymbol{x}'_{i,j+1} - \boldsymbol{x}'_{i,j}  |
    \label{eq:discrete_tv}
\end{equation}

Thus, Eq.~\ref{eq:tv_continuous} can be rewritten in the discrete case as,
\begin{equation}
    \argmin_{\boldsymbol{x}'} \sum_{i,j}  \Big | \boldsymbol{x}'_{i+1,j} - \boldsymbol{x}'_{i,j} \Big | + \Big |\boldsymbol{x}'_{i,j+1} - \boldsymbol{x}'_{i,j}  \Big |  + \frac{\gamma}{2} \Big (\boldsymbol{x}'_{i,j} - \boldsymbol{x}_{i,j} \Big )^2 
    \label{eq:tv_discrete_obj} 
\end{equation}
The optimization problem in Eq.~\ref{eq:tv_discrete_obj} can be solved efficiently utilizing the split-Bregman algorithm for Gaussian noise \cite{getreuer2012rudin}.

\subsection{Aggregation Process}  \label{sec:aggre}
To make better use of multi-resolution signals, we allow images at each scale to communicate with other images that are at their neighboring scales after a denoising process. For a scale, $k$, denote a scale set that lies within the neighborhood of this scale as $\mathcal{N}(k)$. Then, the aggregation process intends to fuse images at scale $k$ with images from $\mathcal{N}(k)$,
\begin{equation}
    \boldsymbol{x}_{(k), a}^t = Agg (\boldsymbol{x}_{(k)}^t, \; \boldsymbol{x}_{(k')}^t ), \; \forall k' \; \in \mathcal{N}(k)
    \label{eq:aggregation}
\end{equation}
where $Agg$ denotes the signal aggregation operator, $\boldsymbol{x}_{(k), a}^t$ denotes an aggregated image after the diffusion-denoising mechanism at step $t$ for a scale $k$, $\boldsymbol{x}_{(k)}^t$ and $\boldsymbol{x}_{(k)'}^t$ can be obtained from Eq.~\ref{eq:diff_denoise}. 

The aggregation process involves two steps. Firstly, we transform an image from a scale $k'$ to scale $k$ by applying $\mathcal{T} (\boldsymbol{x}_{(k)'}^t, k'/k)$ . We then perform an aggregation operation,
\begin{equation}
    \rev{\boldsymbol{x}_{(k), a}^t = \frac{1}{1+ |\mathcal{N}(k)| }  \Big [ \boldsymbol{x}_{(k)}^t + \sum_{k'} \mathcal{T} (\boldsymbol{x}_{(k')}^t, k'/k) \Big ],  \; k' \in \mathcal{N}(k) }
    \label{eq:perform_agg}
\end{equation}
where $|\mathcal{N}(k)|$ denotes the cardinality of the set $\mathcal{N}(k)$. To reduce computation costs, we create $\mathcal{N}(k)$ using the one-hop neighborhood of scale $k$.  

As shown in Fig.~\ref{fig:framework}, the developed diffusive-denoising-aggregation (DDA) block consists of the three processes introduced in Sections \ref{sec:diffusion} to \ref{sec:aggre}, respectively. In the overall framework, we employ several DDA blocks in a serial manner to facilitate the reversing capability of the proposed method. After following $N$ sequential DDA blocks, we perform reverse operations in Eq.~\ref{eq:reverse_mulscale_opt} to convert images at different resolutions to the original scale, then further smooth them by an averaging operation to obtain the final reversed image.

\subsection{Algorithm}
After introducing each component of the proposed framework, this section presents the implementation algorithm of our method. The pseudo-code of the Algorithm is shown in Alg.~\ref{alg:mdda}. 

\begin{algorithm}[ht]
	\footnotesize
	\SetAlgoLined
	\KwData{An adversarial image $\boldsymbol{x}$, an image scale set $\mathcal{K}$, diffusion step $T'$, noise variance $\sigma^2$, regularization parameter $\gamma$.}
 	\KwResult{A reversed image $\boldsymbol{x}_{rev}$ that removes adversarial noises from $\boldsymbol{x}$ using the MDDA framework. }
 	\tcp{Create a multiscale pyramid from $\boldsymbol{x}$} 
 	\For{$k$ in $\mathcal{K}$:}{
 	Compute $\boldsymbol{x}_{(k)}$ using transform in Eq.~\ref{eq:mulscale_opt}\;
 	Initialize $\boldsymbol{x}_{(k)}^0:  \boldsymbol{x}_{(k)}^0 \leftarrow  \boldsymbol{x}_{(k)} $\;
 	}
 	\tcp{Perform DDA operations}
 	\For{$t=1$ \KwTo $T'$ }{
 	\For{$k$ in $\mathcal{K}$}{
 	Generate isotropic Gaussian noises: $\boldsymbol{n} \leftarrow \mathcal{N} \Big( 0, \; \sigma^2/T^{'} \Big)$ \;
 	Update $\boldsymbol{x}_{(k)}^{t}$ by noise injection: $\boldsymbol{x}_{(k)}^{t} \leftarrow  \boldsymbol{x}_{(k)}^{t-1} + \boldsymbol{n} $ \;
 	Update $\boldsymbol{x}_{(k)}^{t}$ by total variation regularization denoising from Eq.~\ref{eq:tv_discrete_obj} using  split Bregman algorithm \;
 	Update $\boldsymbol{x}_{(k)}^{t}$ by projecting $\boldsymbol{x}_{(k)}^{t}$ to image space: $\boldsymbol{x}_{(k)}^{t} \leftarrow \mathcal{P} \Big ( \boldsymbol{x}_{(k)}^{t} \Big )$ \;
 	Update $\boldsymbol{x}_{(k)}^{t}$ by aggregating neighboring resolution images in Eq.~\ref{eq:perform_agg} \;
 	}
 	}
	\tcp{Smooth different scales of images}
	\For{$k$ in $\mathcal{K}$:}{
	Convert $\boldsymbol{x}_{(k)}^{T'}$ to the original scale by Eq.~\ref{eq:reverse_mulscale_opt}: $\boldsymbol{x}_{(k)}^{T'} \leftarrow \mathcal{T}^{-1} (\boldsymbol{x}_{(k)}, k )$ \;
	}
	Compute smoothed image $\boldsymbol{x}_{smooth}$: $\boldsymbol{x}_{smooth} \leftarrow \frac{1}{|\mathcal{K}|} \sum_{i=1}^{|\mathcal{K}|} \boldsymbol{x}_{(i)}^{T'} $  \;
	\textbf{Return}: A reversed image $\boldsymbol{x}_{rev} = \boldsymbol{x}_{smooth}$.
	\caption{The proposed algorithm of MDDA in reversing skin adversarial examples.}
	\label{alg:mdda}
\end{algorithm}

\section{Experimental Results}
\label{sec:exp_results}

In this section, we describe in detail our experimental results including the experimental setup and thorough comparisons to different state-of-the-art baseline methods. 

\subsection{Experimental Setup}

\textbf{Dataset and models:}  We conducted experiments on ISIC 2019, a largest dataset of skin cancer images in the multiclass setting \cite{tschandl2018ham10000, codella2018skin, combalia2019bcn20000}. ISIC 2019 consists of 5331 dermoscopic images, each of which is associated with a diagnosis label as the ground truth. This dataset includes 8 skin diseases: malignant melanoma (MM), melanocytic
nevus (MN), basal cell carcinoma (BCC), actinic keratosis
(AK), benign keratosis (BKL), dermatofibroma (DF), vascular lesion (VASC), and squamous cell carcinoma (SCC). The categories of the diseases and the number of images in that category have been listed in
Table~\ref{tab:dataset_stats}. This dataset can be downloaded at link \footnote{https://challenge.isic-archive.com/landing/2019/}.

\begin{table}[!htbp]
\centering
\caption{Number of images in each of the eight categories in ISIC 2019 dataset: malignant melanoma (MM), melanocytic
nevus (MN), basal cell carcinoma (BCC), actinic keratosis
(AK), benign keratosis (BKL), dermatofibroma (DF), vascular lesion (VASC), and squamous cell carcinoma (SCC). }
\label{tab:dataset_stats}
\begin{adjustbox}{width=0.48\textwidth}
\begin{tabular}{cccccccc}
\toprule
 {MM}	& {MN}  & {BCC} & {AK} & {BKL} & {DF} & {VASC} & {SCC} \\
 4522  & 12875 & 3323 & 867 & 2624 & 239 & 253 & 628 \\ 
 \bottomrule
\end{tabular}
\end{adjustbox}
\end{table}

We utilize two commonly used DCNN models for skin cancer diagnosis: ResNet50 \cite{ResNet} and MobileNetV2 \cite{MobileNetV2}. Two models were trained using the code provided in work \cite{wang2022ssd} on a training/validation/test dataset split as 8:1:1. For both models, their input images are resized to be $224 \times 224$.

\textbf{Attacks and defense baselines:} Given a pretrained skin diagnostic model, we assume an attacker has white-box access to the model. Namely, the attacker has a full knowledge of the victim model, including the model weights and training setups. Then, an attacker can perform test-time attacks on the victim model using different types of attacks. To be specific, this paper adopts six popular adversarial attacks: FGSM \cite{FGSM}, BIM \cite{Iteradv17}, CW \cite{CW17}, PGD \cite{madry2018towards}, DIFGSM \cite{transfer19} and AutoAttack \cite{croce2020reliable}. 

To protect victim skin diagnostic models, model owners will select and evaluate the effectiveness of defense methods. For comparison, we adopt three typical defenses as our baseline methods: bit depth reduction (BDR) \cite{Xu0Q18}, super-resolution (SR) \cite{mustafa2019image} and neural representation purifier (NRP) \cite{naseer2020self}. For a fair comparison, we utilize the default parameters in their respective officially released implementations in the following experiments.

\textbf{Parameters and metrics:} In the attack setting, we set the perturbation budget as $\epsilon=2/255$ for FGSM \cite{FGSM}, BIM \cite{Iteradv17}, PGD \cite{madry2018towards}, DIFGSM \cite{transfer19} and AutoAttack \cite{croce2020reliable}, which can almost fool victim models completely in the white-box setting. In addition, we also conducted experiments with a large perturbation budget, i.e., $\epsilon=6/255$, yet with acceptable visual quality of generated adversarial examples. For CW \cite{CW17} attack, we used a learning rate as 0.01. For all attacks, we set their iteration steps as 100 to ensure all attacks can achieve high attack success rates. 

During evaluations, we adopt classification accuracy as the metric. An attack method will decrease the classification accuracy significantly, while a good defense method will help increase the classification accuracy of a model that is being attacked. Therefore, a higher classification accuracy indicates a stronger protective capability of a defense method. Such a defense method is thus preferred.

\subsection{Comparison to Baseline Defenses}
In this section, we evaluate and compare different defense methods with the victim model as ResNet50 pretrained on ISIC 2019 dataset under the white-box non-targeted attack setting. \rev{In such an attack setting, an attacker is assumed to have full access to the victim ResNet50 model and can directly craft adversarial examples based on the victim model. We report the performance comparison results ($\%$) of different defense methods in Table~\ref{tab:resnet_50_eps2} for $\epsilon=2/255$ and $\epsilon=6/255$ under the $\ell_{\infty}$ norm for FGSM, BIM, PGD, DIFGSM, and AutoAttack attacks in the white-box setting.} In the proposed method, we set the noise variance parameter $\sigma^2$ as 0.125, and the number of DDA blocks as 5 and 3 for $\epsilon=2/255$ and $\epsilon=6/255$, respectively.  

\begin{table*}[!htbp]
\caption{\rev{Performance comparison of different defense methods on the ISIC 2019 dataset under white-box non-targeted attacks on ResNet50 model. Each entry shows the model accuracy (\%) under a defense and attack combination pair. The first column denotes different defense methods. Columns 2 to 8 show accuracy results under the settings of clean samples and adversarial samples by 6 white-box attacks, respectively. The perturbation budget is $\epsilon=2/255$ and $\epsilon=6/255$ for FGSM, BIM, PGD, DIFGSM and AutoAttack methods under the $\ell_{\infty}$ norm. Best performances are marked in bold. }}
 
 \label{tab:resnet_50_eps2}
    \centering
\begin{adjustbox}{width=0.9\textwidth}
\begin{tabular}{ccccccccc}
    \toprule
        Defenses & Clean & FGSM & BIM & CW & PGD & DIFGSM & AutoAttack & $\overline{\textrm{ACC}}$  \\ \hline
        No defense & \textbf{82.0} & 0.99/12.55 & 0/0 & 0 & 0/0 & 0.04/0 & 0/0 & 11.9/13.5  \\ \hline
        BDR  & 53.8 & 6.5/10.3 & 5.9/0 & 17 & 9/0 & 5.8/0 & 9.5/3 & 15.4/12.0 \\ \hline
        SR & 79.1 & 6.2/7.5 & 14.4/0 & 18.6 & 18.3/0.1 & 7.4/0 & 22.8/3.3 & 23.8/15.5  \\ \hline
        NRP & 52.8 & 30.6/\textbf{65.6} & 37.3/30.3 & 39.6 & 38.5/32.3 & 34.9/24.4 & 39.1/33.2 & 39/39.7  \\ \hline
        \textbf{MDDA} & 69.1/65.7 & \textbf{56.8}/46.1 & \textbf{62.3}/\textbf{53.7} & \textbf{63.2}/\textbf{61} & \textbf{62.3}/\textbf{53} & \textbf{59.7}/\textbf{47.8} & \textbf{64.3}/\textbf{53.6} & \textbf{62.5}/\textbf{54.4}  \\
        \bottomrule
\end{tabular}
\end{adjustbox}
\end{table*}

In Table~\ref{tab:resnet_50_eps2}, we observe that the victim model can achieve high accuracy (i.e., 82\%) without being attacked. However, as shown in the second row (i.e., ``No defense'') the classification accuracy can be significantly reduced without applying defense methods. Among the attack methods, we observe that FGSM performs worse than other attacks since it cannot fool ResNet50 completely even without using any defenses. This is because FGSM is obtained based on a linearity assumption, and it can ``underfit'' a DCNN model \cite{MIFGSM}. Therefore, we mainly focus on other typical attacks during the defense evaluation.

Among the defense methods, the proposed method generally achieves the best defensive capability among different attacks. Notably, for DIFGSM attack, the strongest defense method, our method outperforms the second best defense method by \textbf{24.8}\% and \textbf{23.4}\% for $\epsilon=2/255$ and $\epsilon=6/255$ settings, respectively. More importantly, we observe that the proposed MDDA method performs similarly well regardless of the types of adversarial attacks. This interesting observation indicates that our method satisfies the \textbf{attack-agnostic} property.

\begin{figure*}[!htbp]
     \centering
     \captionsetup[subfigure]{justification=centering}  
     \begin{subfigure}[b]{0.45\textwidth}
         \centering
         \includegraphics[width=\textwidth]{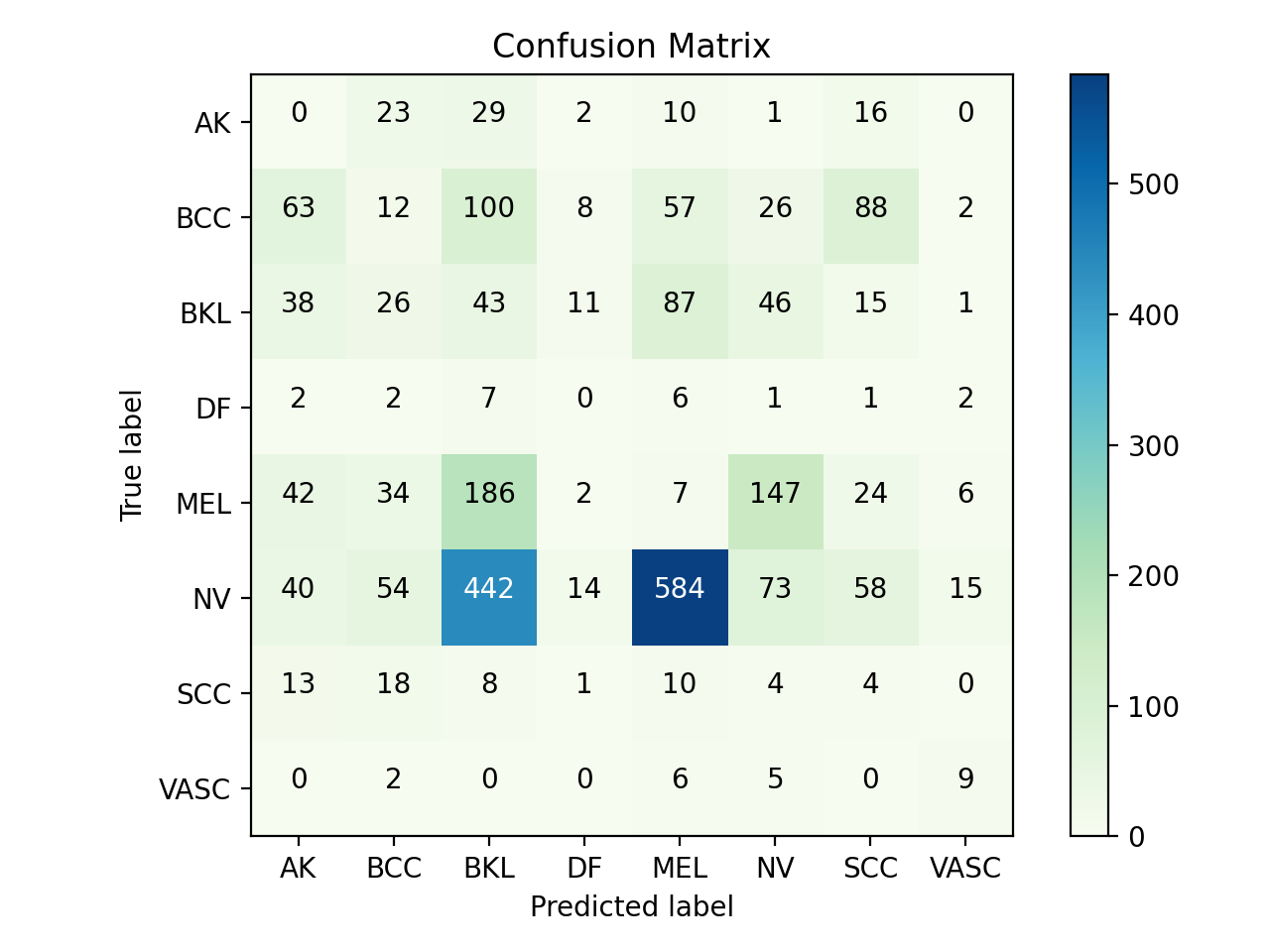}
         \caption{BDR}
         \label{fig:bdr_eps2}
     \end{subfigure}
     \begin{subfigure}[b]{0.45\textwidth}
         \centering
         \includegraphics[width=\textwidth]{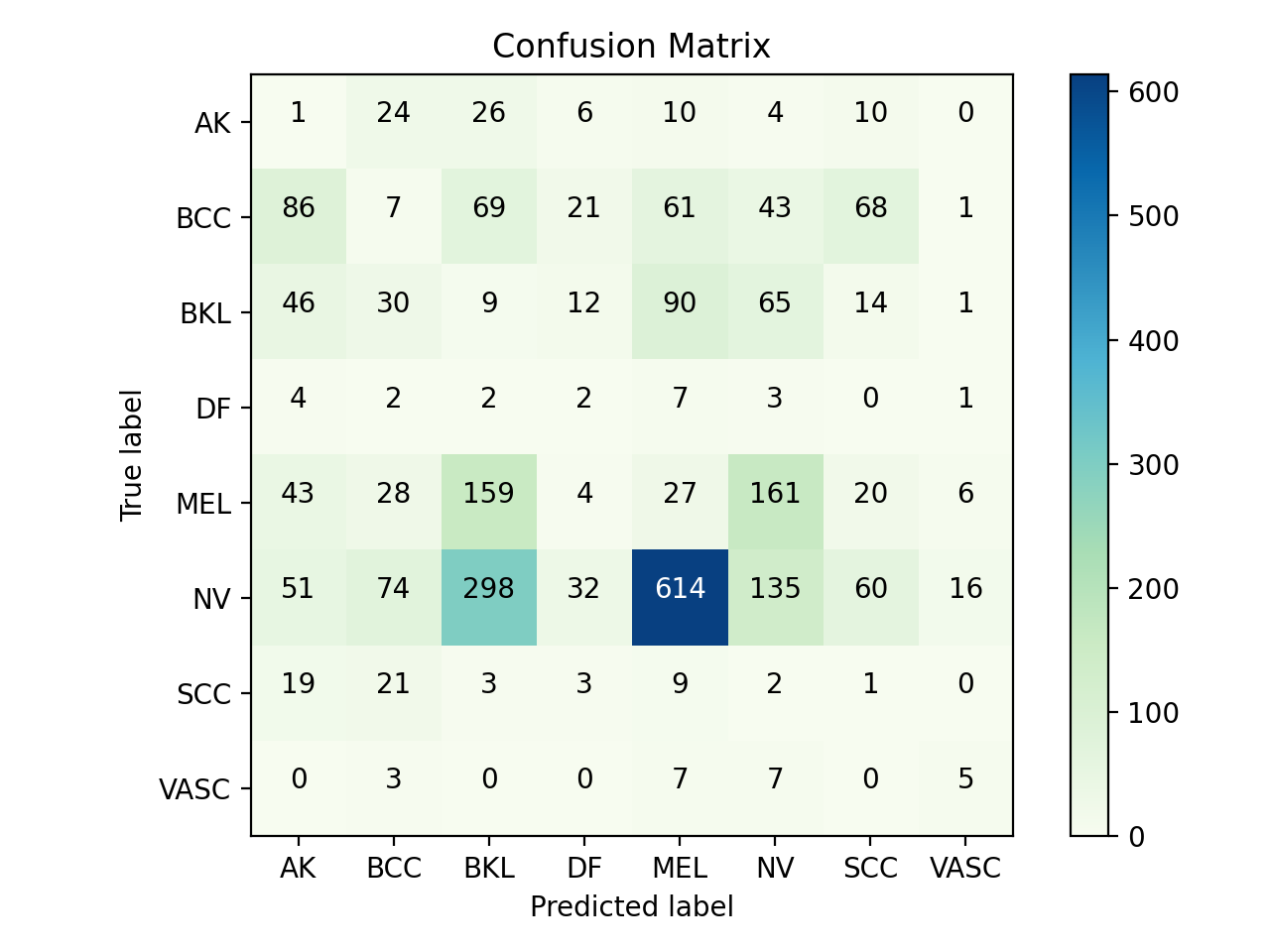}
         \caption{SR}
         \label{fig:sr_eps2}
     \end{subfigure}
    \\[4mm]
     \begin{subfigure}[b]{0.45\textwidth}
         \centering
         \includegraphics[width=\textwidth]{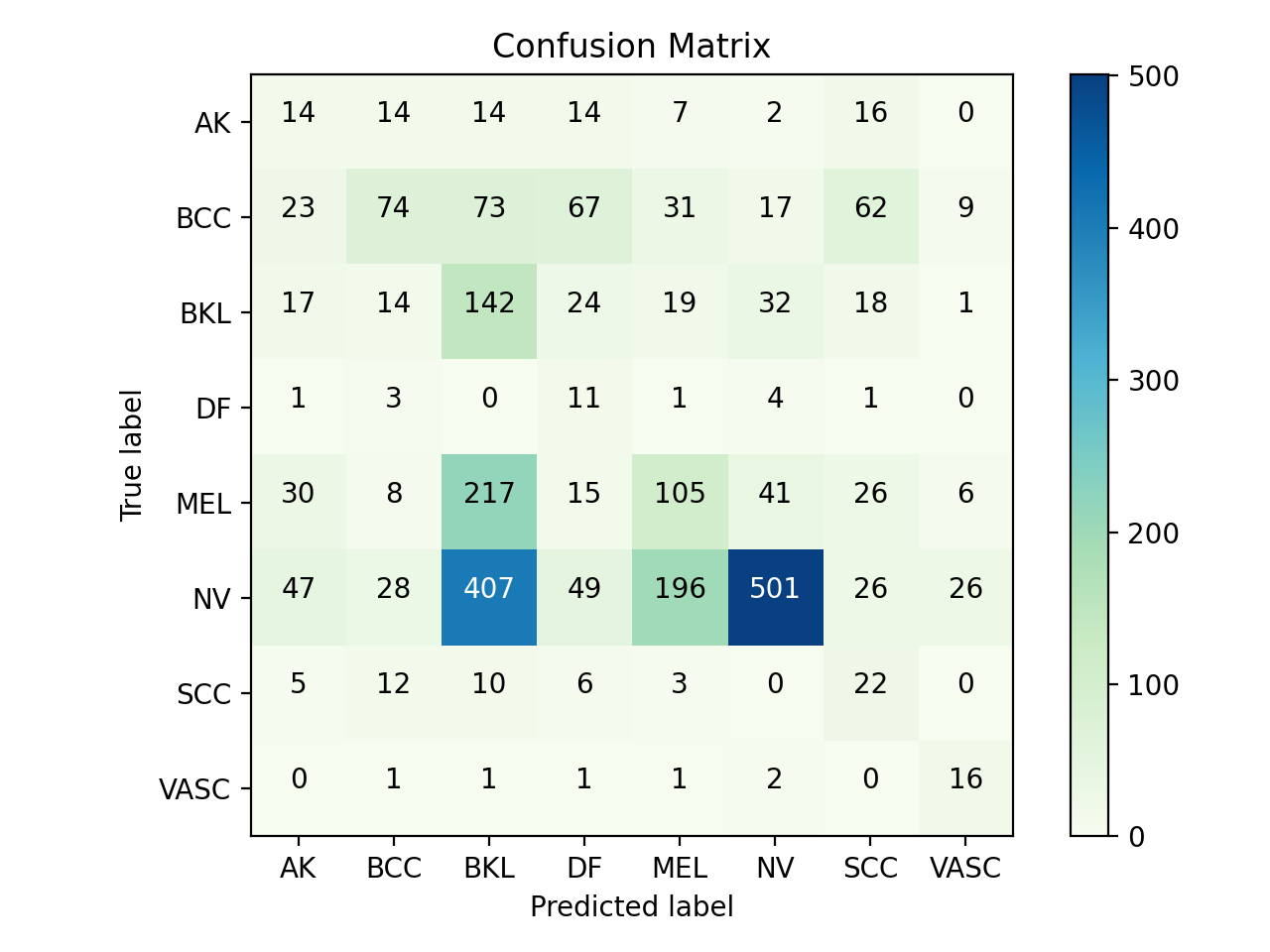}
         \caption{NRP}
         \label{fig:nrp_eps2}
     \end{subfigure}
     \begin{subfigure}[b]{0.45\textwidth}
         \centering
         \includegraphics[width=\textwidth]{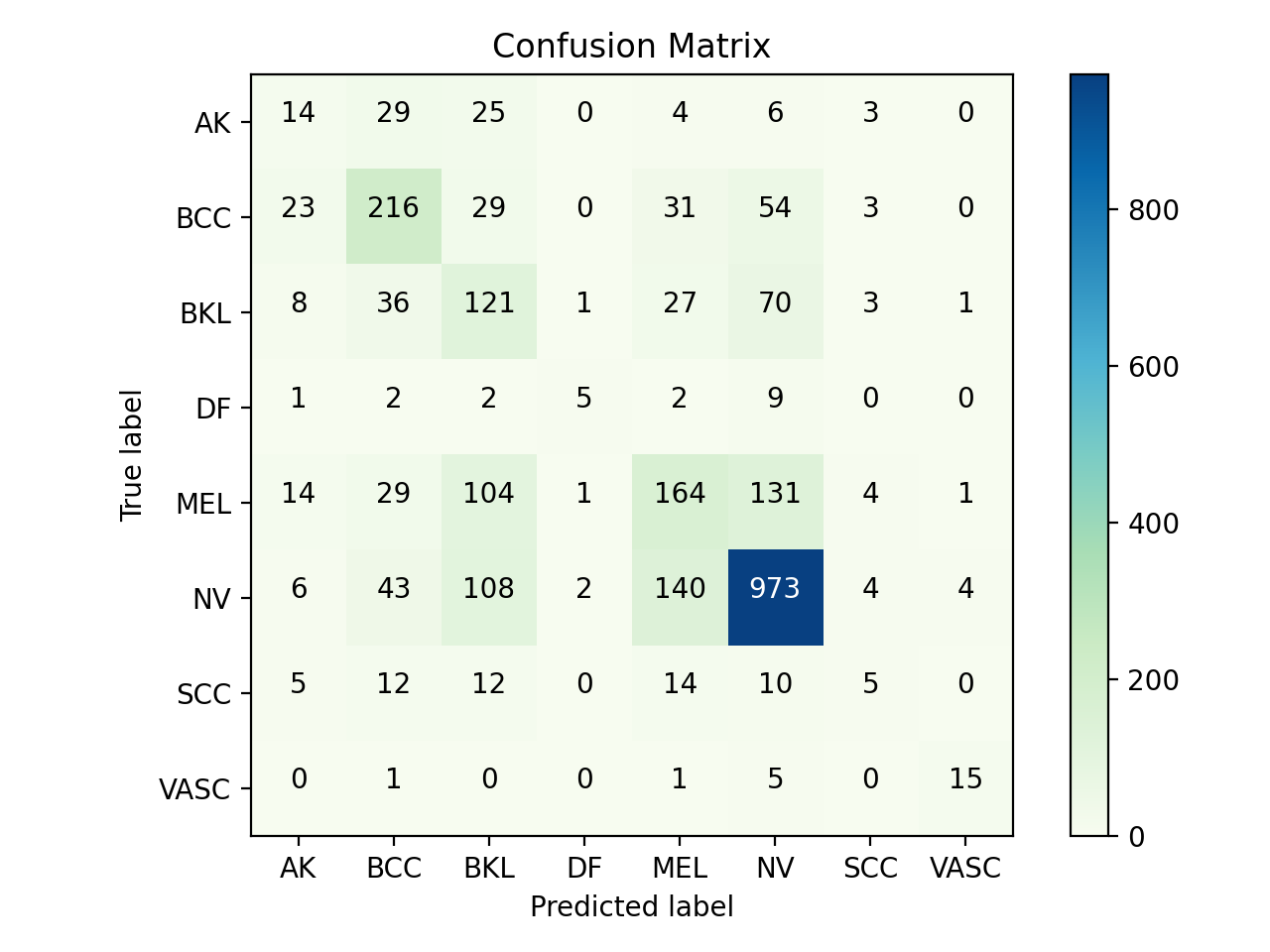}
         \caption{MDDA}
         \label{fig:mdda_eps2}
     \end{subfigure}     
        \caption{Visualizations of confusion matrices from the ResNet50 model for the prediction ability under DIFGSM attack ($\epsilon=2/255$) with four defense methods: (a) BDR, (b) SR, (c) NRP and (d) MDDA.}
        \label{fig:conf_mat_eps2}
\end{figure*}

\begin{figure*}[!htbp]
     \centering
     \captionsetup[subfigure]{justification=centering}  
     \begin{subfigure}[b]{0.45\textwidth}
         \centering
         \includegraphics[width=\textwidth]{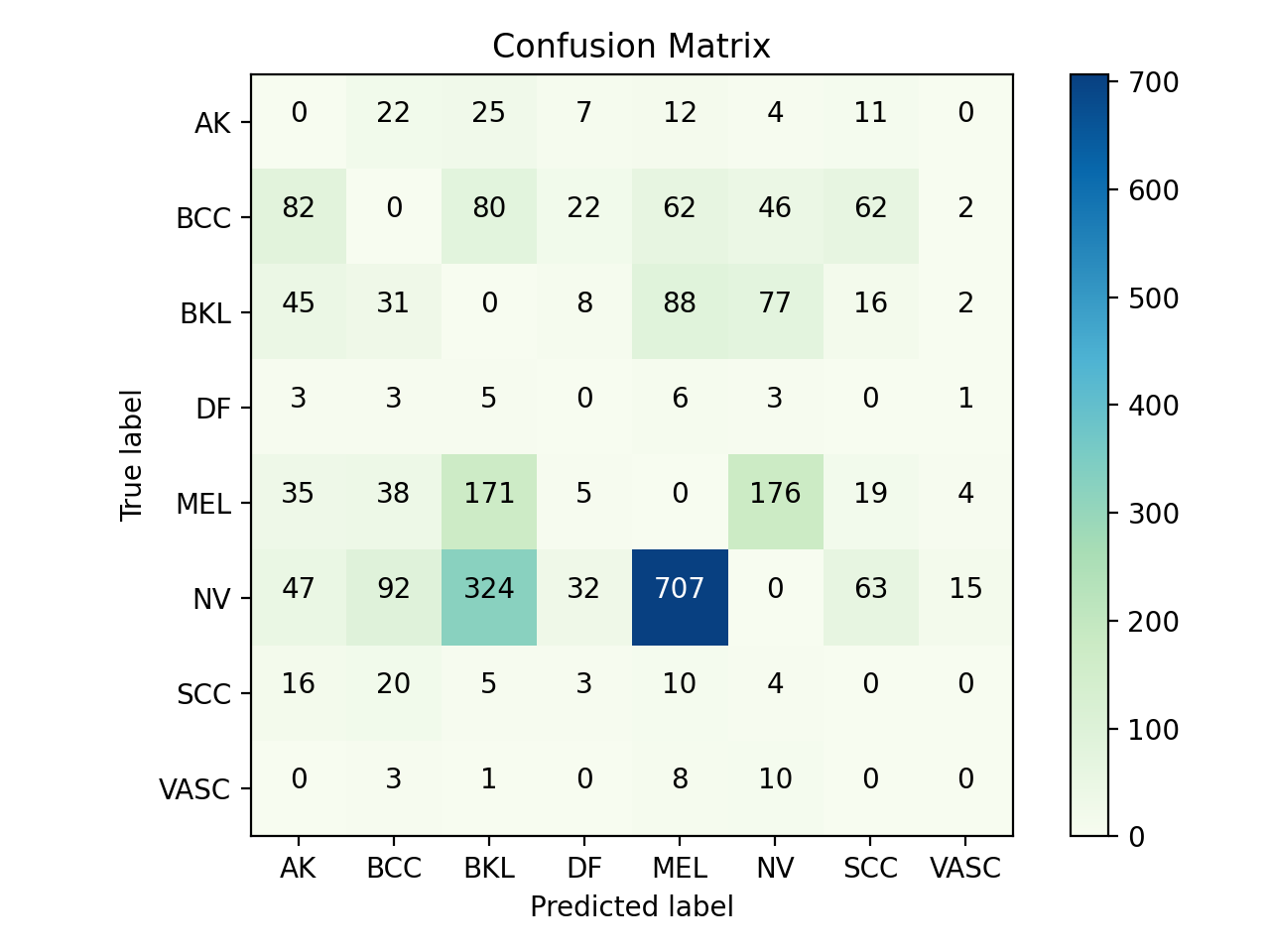}
         \caption{BDR}
         \label{fig:bdr_eps6}
     \end{subfigure}
     \begin{subfigure}[b]{0.45\textwidth}
         \centering
         \includegraphics[width=\textwidth]{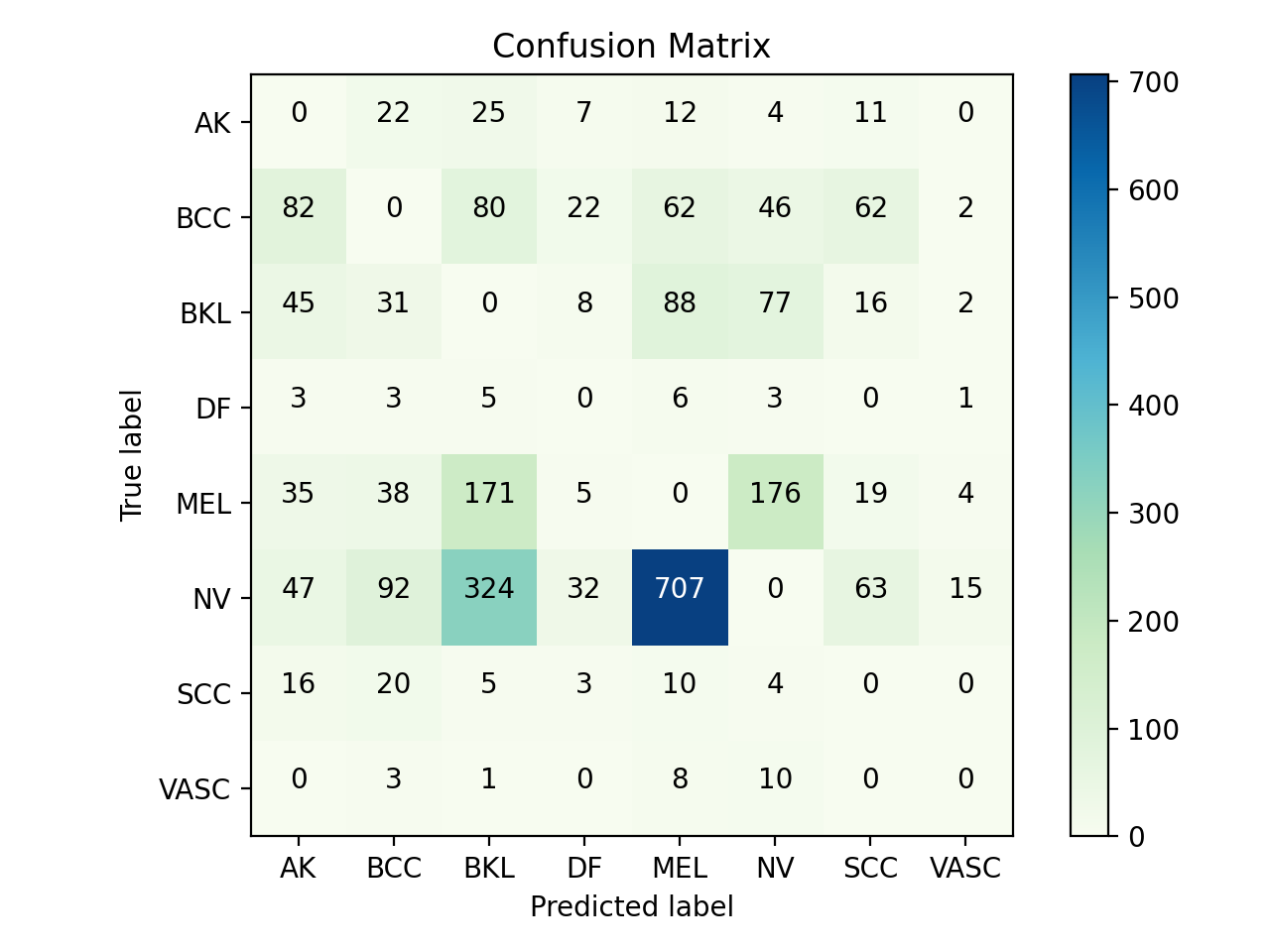}
         \caption{SR}
         \label{fig:sr_eps6}
     \end{subfigure}
    \\[4mm]
     \begin{subfigure}[b]{0.45\textwidth}
         \centering
         \includegraphics[width=\textwidth]{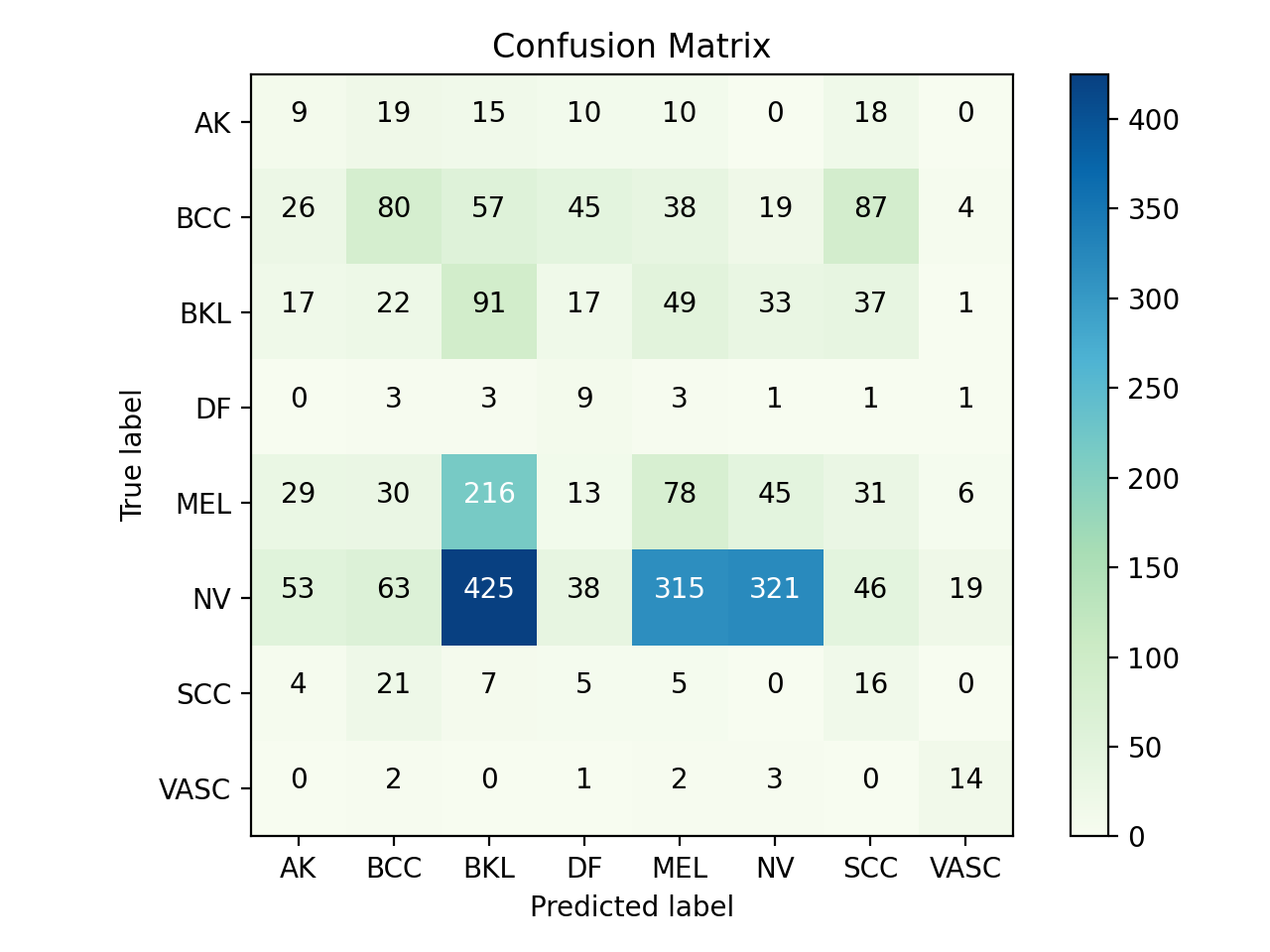}
         \caption{NRP}
         \label{fig:nrp_eps6}
     \end{subfigure}
     \begin{subfigure}[b]{0.45\textwidth}
         \centering
         \includegraphics[width=\textwidth]{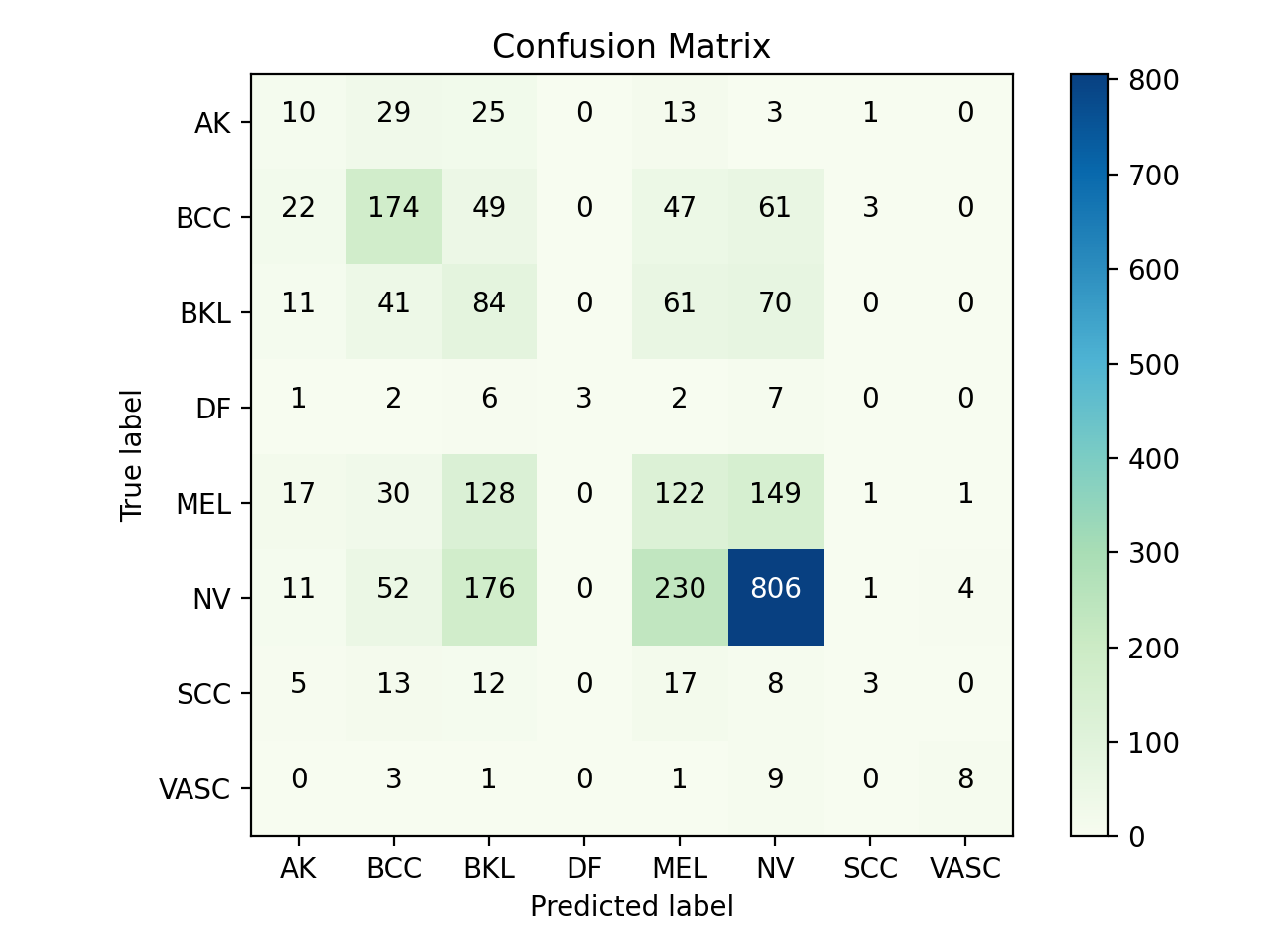}
         \caption{MDDA}
         \label{fig:mdda_eps6}
     \end{subfigure}     
        \caption{\rev{Visualizations of confusion matrices from the ResNet50 model for the prediction ability under DIFGSM attack ($\epsilon=6/255$) with four defense methods: (a) BDR, (b) SR, (c) NRP and (d) MDDA.}}
        \label{fig:conf_mat_eps6}  
\end{figure*}

We next zoom in on prediction classes and compare the protection of each defense method to each of the eight individual categories. Based on the analysis above, we choose DIFGSM attack since it poses the most dangerous threat to the victim ResNet50 model and it has the highest resistance to adversarial defense methods.  In Fig.~\ref{fig:conf_mat_eps2} and Fig.~\ref{fig:conf_mat_eps6}, we visualize the confusion matrices from ResNet50 regarding the classification performance under DIFGSM attack with the perturbation budgets as $\epsilon=2/255$ and $\epsilon=6/255$, respectively. \rev{The experimental results reveal that, compared with baseline methods BDR and SR, the proposed MDDA method achieves superior prediction performances for each of the eight skin lesion classes. Compared with NRP, our method surpasses or achieves comparable performance with it for classes BCC, MEL, NV, and AK. Yet, MDDA is outperformed by NRP for smaller-sample classes BKL, DF, SCC and VASC. A possible reason for the inferior performance of MDDA on smaller-size samples could be that the pretrained victim model was not immune enough to noise, particularly for the smaller-size samples. Indeed, even on clean data, the recognition accuracy of these samples performs worse than other classes for the victim model, and the additive noise injected by our defense may further deteriorate the classification accuracy.  }

\subsection{Cross-architectural Protection}
In addition to evaluating the defensive performances in the white-box setting, we also conducted experiments to compare defense methods in the cross-architectural model protection scenes. \rev{Different from the white-box attack setting, an attacker cannot access the victim model. Instead, the attacker can employ a substitute model (usually with different architectures from the victim one) to generate adversarial examples. Indeed, this is a more practical attack/defense scenario since an attacker may not have full access to a victim model (i.e., MobileNetV2). A white-box attack can then turn into a cross-architectural attack utilizing a different victim model.} Thus, he/she may craft skin adversarial examples based on a well-trained ResNet50, in the hope of fooling MobileNetV2 successfully. 

Under this more realistic attack setting, we evaluate the defensive capability of each of the four defense methods. We select four strong attacks, namely BIM, PGD, DIFGSM, and AutoAttack. We choose a large perturbation ball $\epsilon=6/255$. The results are reported in Table~\ref{tab:mobilenet_eps6}. 

From this table, we can observe that even for the ``No defense'' case, the attack ability of each attack method actually reduced significantly compared to that of ResNet50 model. For example, the averaged accuracy is 40.8\% in MobilenetV2 (first row in Table~\ref{tab:mobilenet_eps6}), while it is 0 for the same four attack methods. The reason is that MobileNetV2 has a quite different network architecture from ResNet50, thus limiting the attack ability of each attack method. As for defense methods, we observe the improvements of averaged accuracy of SR, NRP, and MDDA over ``No defense'' are: 6.5\%, 3.7\%, 16.9\%, respectively. Since BDR has almost no improvement in average accuracy, we omit this defense in this section.  It is apparent that our MDDA method can still maintain high defensive effectiveness over the baseline methods under the cross-architectural protection situation. 

\begin{figure*}[!htbp]
     \centering
     \captionsetup[subfigure]{justification=centering}  
     \begin{subfigure}[b]{0.45\textwidth}
         \centering
         \includegraphics[width=\textwidth]{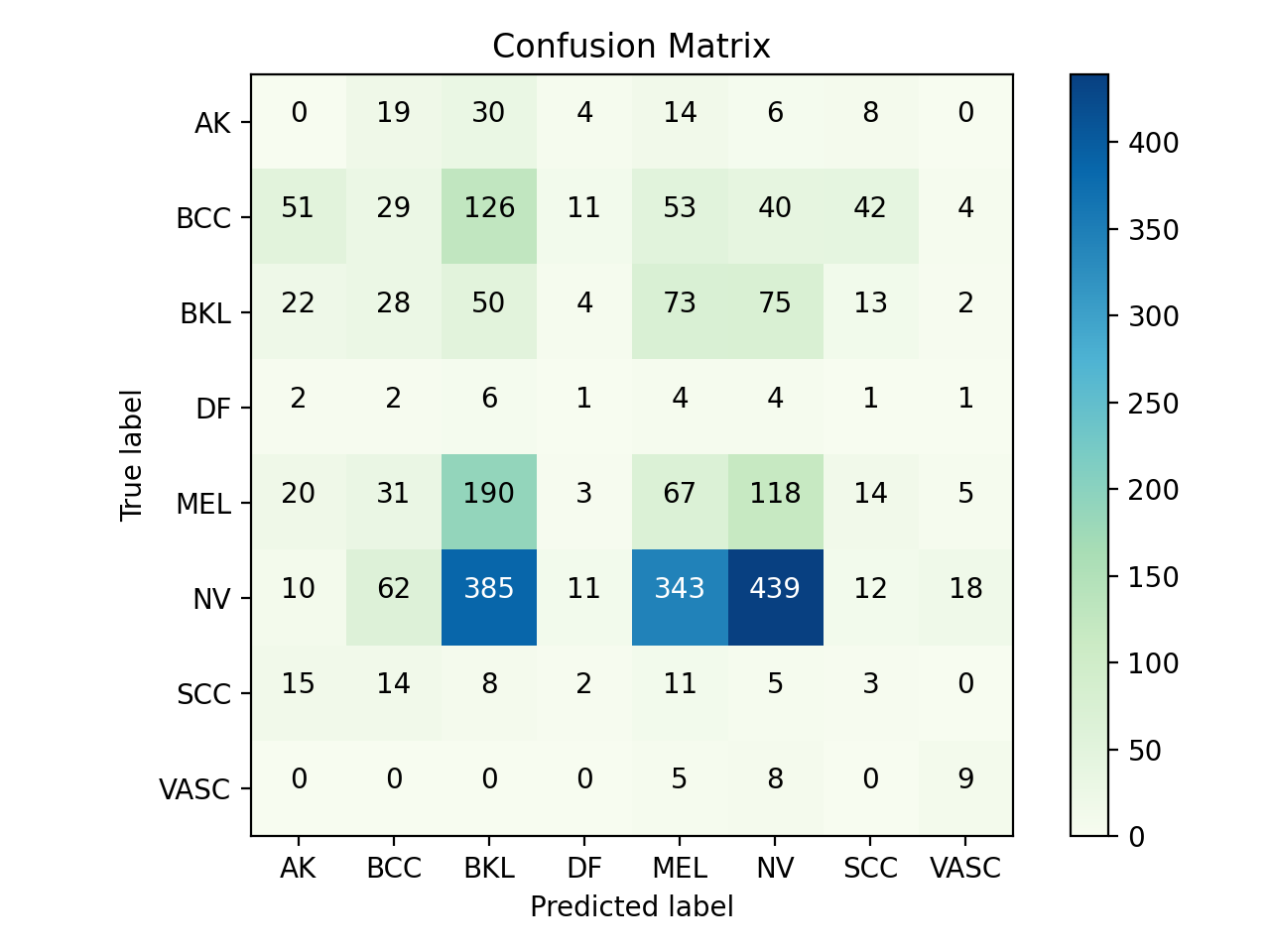}
         \caption{No defense}
         \label{fig:mobilenet_bdr_eps6}
     \end{subfigure}
     \begin{subfigure}[b]{0.45\textwidth}
         \centering
         \includegraphics[width=\textwidth]{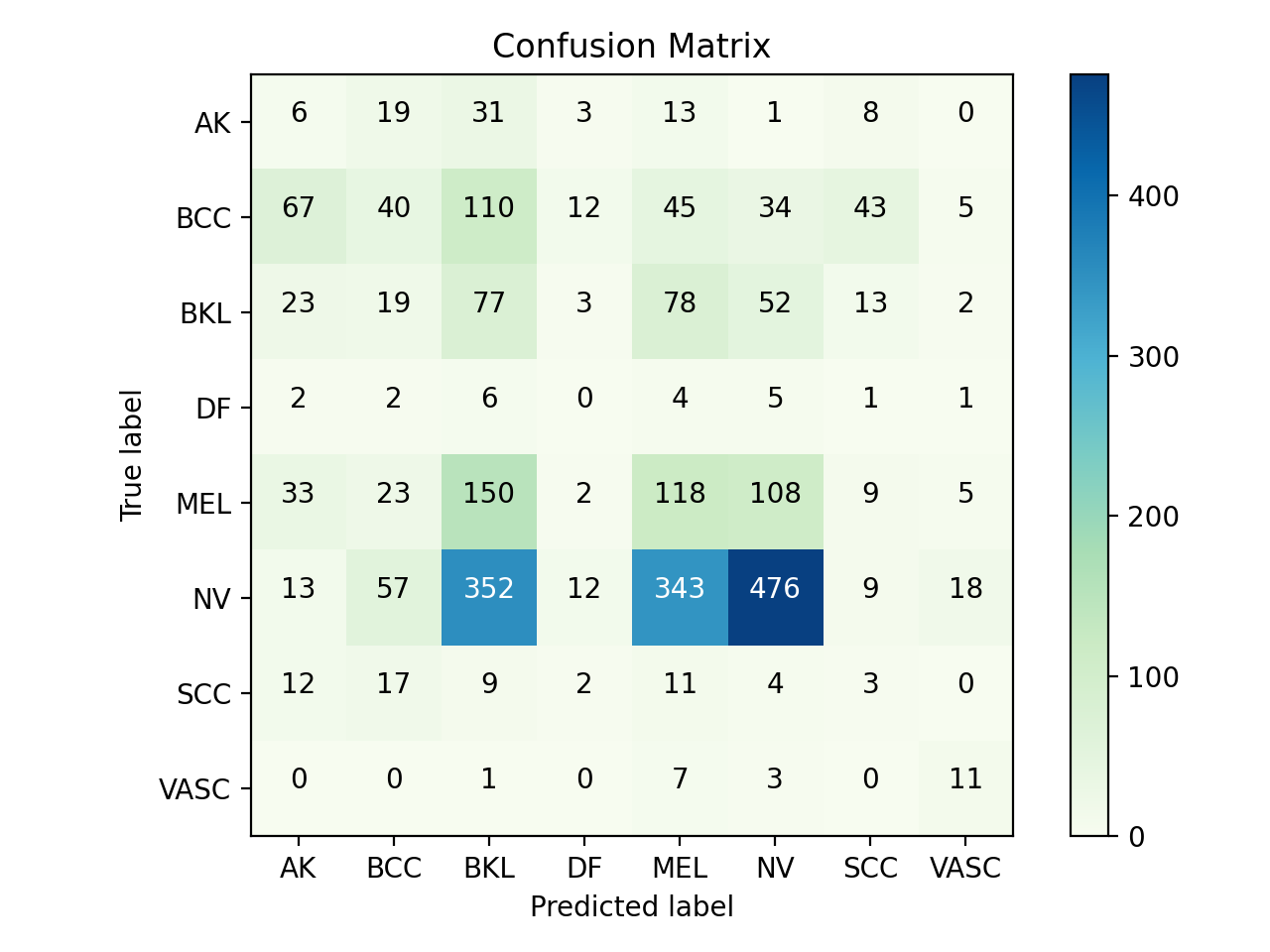}
         \caption{SR}
         \label{fig:mobilenet_sr_eps6}
     \end{subfigure}
    \\[4mm]
     \begin{subfigure}[b]{0.45\textwidth}
         \centering
         \includegraphics[width=\textwidth]{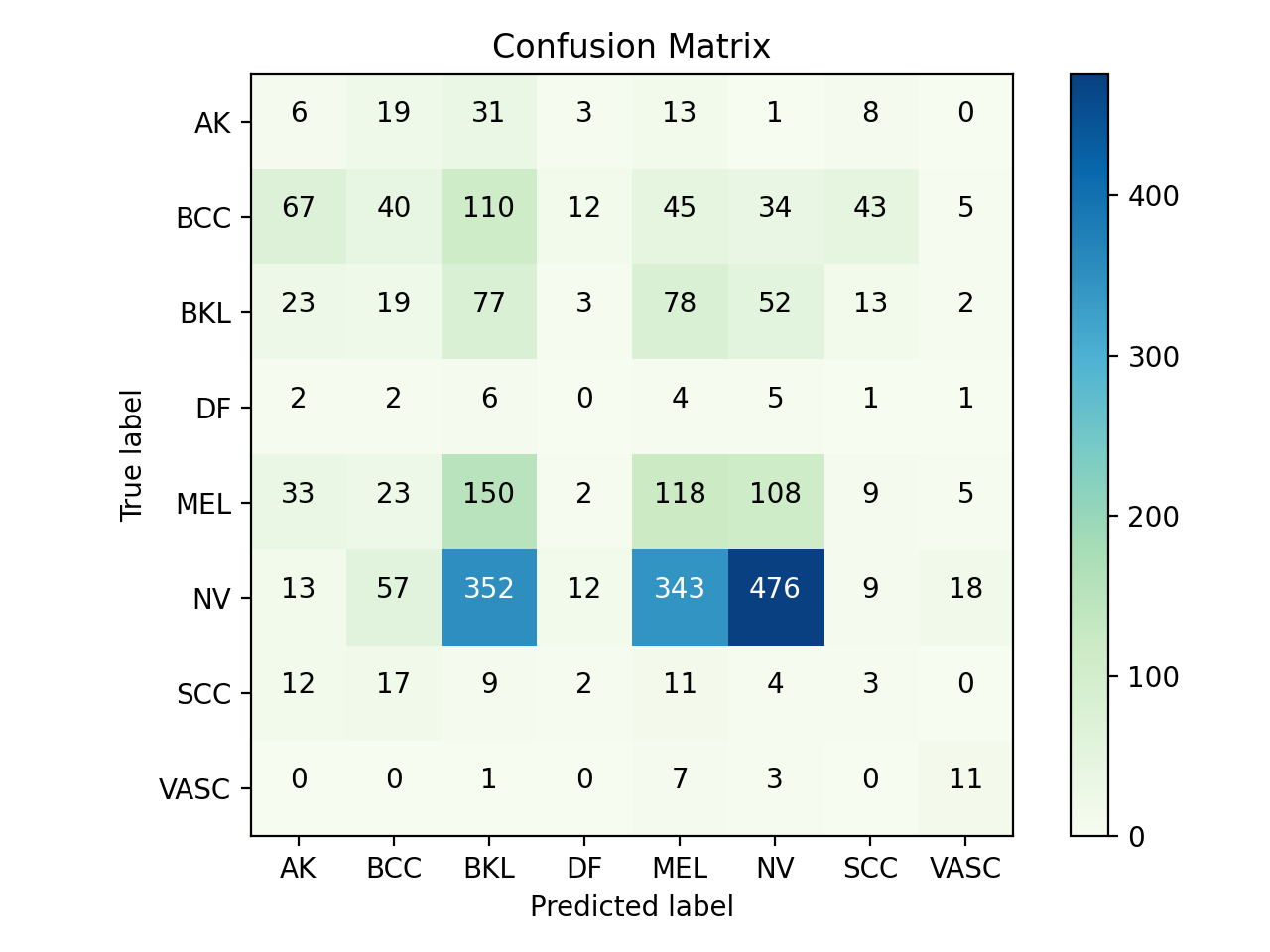}
         \caption{NRP}
         \label{fig:mobilenet_nrp_eps6}
     \end{subfigure}
     \begin{subfigure}[b]{0.45\textwidth}
         \centering
         \includegraphics[width=\textwidth]{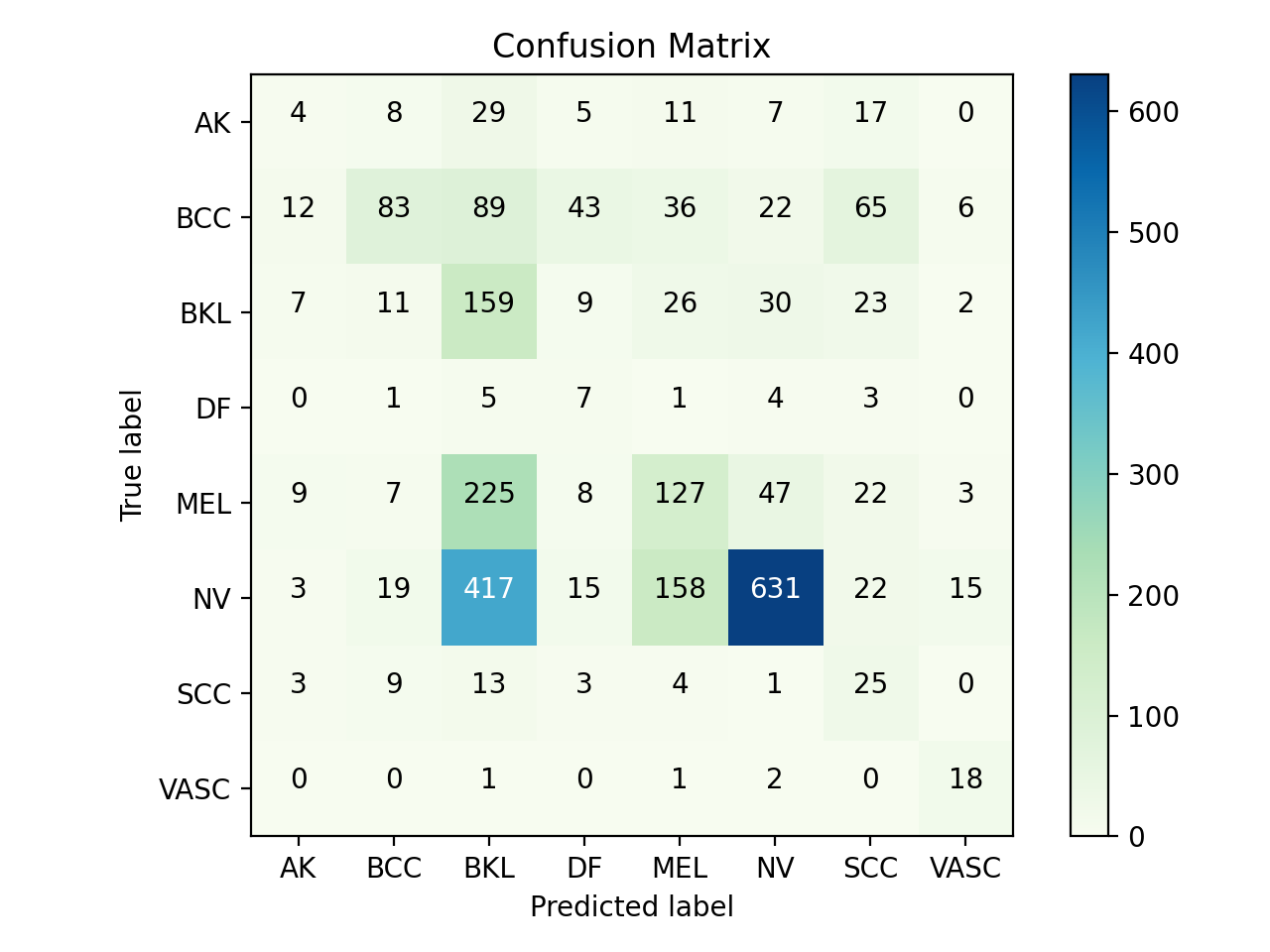}
         \caption{MDDA}
         \label{fig:mobilenet_mdda_eps6}
     \end{subfigure}     
        \caption{Visualizations of confusion matrices from the MobileNetV2 model for the prediction ability under DIFGSM attack ($\epsilon=6/255$) with (a) No defense, and three defense methods: (b) SR, (c) NRP and (d) MDDA.}
        \label{fig:mobilenet_conf_mat_eps6}
\end{figure*}

We also visualize the confusion matrices for each class in Fig.~\ref{fig:mobilenet_conf_mat_eps6} for a detailed performance comparison. We observe that MDDA is effective for each class in defending against cross-architectural attacks, and it can outperform both SR and NRP in a majority of skin lesion classes. 

\rev{From the defense experiments above, we can observe that our method is generally applicable to defend against both white-box attacks and cross-architectural attacks as we do not have assumptions about the victim model.  Besides, the experiments reveal that the diffusive and denoising processes can effectively reverse the adversarial noises from adversarial examples, and the aggregation procedure further purifies these noises. Regardless of how these adversarial examples were generated since these examples have been reverted to normal samples, they should perform well on the victim model. }

\begin{table}[!htbp]
\caption{\rev{Cross-architectural performance comparison of different defense methods on the ISIC 2019 dataset under cross-architectural non-targeted attacks on MobileNetV2 model. Each entry shows the model accuracy (\%) under a defense and attack combination pair. The perturbation budget is $\epsilon=6/255$ for BIM, PGD, DIFGSM, and AutoAttack methods under the $\ell_{\infty}$ norm. Best performances are marked in bold. }
}
\label{tab:mobilenet_eps6}
    \centering
    \begin{adjustbox}{width=0.48\textwidth}
    \begin{tabular}{cccccc}
    \toprule
        Defense method & Clean & BIM & DIFGSM & AutoAttack & $\overline{\textrm{ACC}}$  \\ \hline
        No defense & \textbf{75.4} & 33.1 & 23.6 & 31.2 & 40.8  \\ \hline
        SR & 71 & 42.3 & 28.9 & 43 & 46.3  \\ \hline
        NRP & 46.7 & 44.2 & 41.6 & 44.7 & 44.3  \\ \hline
        \textbf{MDDA} & 60.6 & \textbf{56.8} & \textbf{55.7} & \textbf{57.7} & \textbf{57.7}  \\ 
        \bottomrule
		\end{tabular}
		\end{adjustbox}
\end{table}

\subsection{Parameter Sensitivity}
In this section, we study the sensitivity of parameters in the proposed method. There are two important parameters in MDDA: the noise variance $\sigma^2$ and the number of DDA blocks $N$. \rev{We select $\sigma^2$ using grid search with a range from 0.05 to 0.175 with a stepsize as 0.025 on the validation set} and we fix $N=3$ in the $\epsilon=6/255$ setting. We observe that as the noise variance increases, the classification accuracy on the clean data decreases, meanwhile the defense capability increases. Considering we may not be able to distinguish adversarial examples from clean ones, the accuracy of clean samples should not be too low. We set a threshold for clean accuracy as 65\%, which is comparable to clinical doctors in multiclass skin disease classification. In this way, we can choose $\sigma^2=0.125$ with its clean accuracy as 65.7\%. For consistency, we also adopt $\sigma^2=0.125$ for the $\epsilon=2/255$ setting, yet we believe a more careful parameter tuning can further improve our method's performance. 

\begin{figure}[!htbp]
\centering
\includegraphics[width=0.48\textwidth]{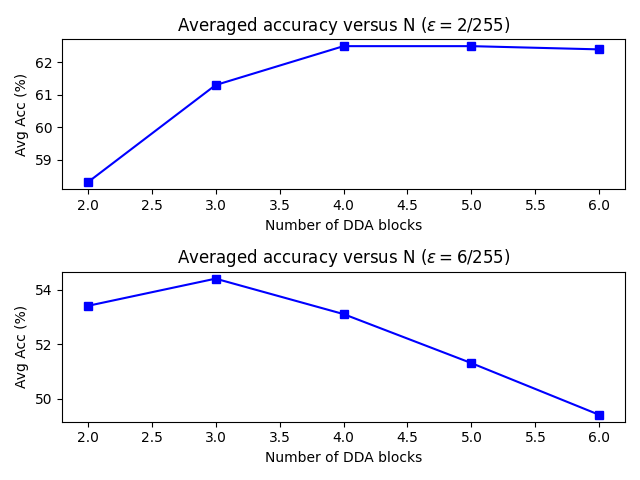}
\caption{The sensitivity curve of averaged accuracy over the number of DDA blocks in the MDDA defense. }
\label{fig:para_sensitivity}
\end{figure}

For the selection of $N$, we grid search an optimum value specific to $\epsilon=2/255$ and $\epsilon=6/255$ settings. As shown in Fig.~\ref{fig:para_sensitivity}, we can observe that \rev{the impact of $N$ is relatively stable when the number of blocks is larger than 4 for the $\epsilon=2/255$ setting}. Yet the protection performance tends to be more influenced by $N$ when defending against heavily perturbed adversarial examples (e.g., $\epsilon=6/255$). \rev{An intuitive explanation for this phenomenon is that the injected Gaussian noise cannot effectively destroy the structure of adversarial noises as $N$ increases, yet the redundant additive noise from the denoising procedure may still hinder the model from correctly recognizing the noisy samples, thus reducing the averaged accuracy.} In this case, we suggest adopting a small number of DDA blocks to guarantee defensive capability. \rev{It is worth mentioning that MDDA can achieve significantly higher averaged accuracy than three baselines with $N=3$ for both settings $\epsilon=2/255$ and $\epsilon=6/255$. This indicates the attack-agnostic property of the proposed method.}

\section{Discussion}
\label{sec:discussion}
In the previous section, we experimentally demonstrate the effectiveness of the proposed MDDA in defending against different kinds of adversarial attacks on skin diagnostic models. Although MDDA can outperform state-of-the-art defenses by a large margin, it still faces several challenges and reminds to be further improved. For example, our noise injection can also degrade the recognition accuracy of clean images, i.e., 69.3\% (ours for $\epsilon=2/255$) vs 82.0\% (clean accuracy). \rev{To resolve this issue, a promising approach is to first employ adversarial example detectors (e.g. \cite{roth2019odds,liu2019detection}) that distinguish clean samples from adversarial ones. Then we can adopt reconstruction-based defenses (e.g. MDDA) to purify the detected adversarial examples only. In this way, we may further improve the average accuracy. }

Besides, despite the fact that MDDA achieves high overall classification accuracy, there still has much space to improve the protection ability for some specific skin lesion classes (e.g., actinic keratosis or AK). \rev{A possible reason is that the original model itself shows different recognition performances for different skin cancer classes. For example, for both ResNet50 and MoblileNetV2 models, the precision of class AK is much lower than class NV, indicating the difficulty of classifying AK even without perturbations to the original image samples. Moreover, since our method requires the injection of Gaussian noises, it becomes even more challenging to correctly classify the hard classes such as class AK under noisy conditions. We hypothesize that applying data augmentation strategies (e.g., augmentation by noise injection) for the original models’ training process could enhance the protection ability of our method, and we will leave it as our future work. }

\section{Conclusion  and Future Work}
In this work, we aim to protect skin diagnostic robots from adversarial attacks and assist in reliably recognizing the type of skin cancer diseases. Specifically, we explore the feasibility of reversing skin adversarial examples as an effective defense method. To fight against adversarial noises, we develop a framework named MDDA that can satisfy attack-agnostic, training-free, and resource-efficient properties simultaneously. MDDA gradually injects Gaussian noises to disrupt the structure of adversarial noises. We also specifically employ the multiscale processing technique to better preserve the key structures of skin images. Importantly, we design a DDA block that can effectively reverse adversarial noises and suppress additive-injected noises. Experimental results on a real skin image dataset show that MDDA can successfully reverse skin adversarial examples and it outperforms state-of-the-art defense methods in defending against white-box attacks and cross-architectural attacks. \rev{In the future, we plan to evaluate our defense on other medical imaging applications and datasets. }

\bibliographystyle{elsarticle-num} 
\bibliography{defense}

\end{document}